\documentclass[acmsmall]{acmart}

\usepackage{threeparttable}
\usepackage{lineno,hyperref}
\usepackage{mathtools,amsthm}
\usepackage[ruled,vlined]{algorithm2e}
\SetKwInput{kwServer}{Server executes}
\SetKwInput{Output}{Output}
\SetKwProg{Fn}{ClientUpdate}{}{}
\SetKwProg{Inference}{Inference}{}{}
\usepackage{multirow}
\usepackage{booktabs}
\usepackage{soul}
\usepackage{xcolor}
\usepackage{bm}

\DeclareMathOperator*{\argmin}{arg\,min}

\def\BibTeX{{\rm B\kern-.05em{\sc i\kern-.025em b}\kern-.08emT\kern-.1667em\lower.7ex\hbox{E}\kern-.125emX}}
    
\copyrightyear{2018}
\acmYear{2018}
\setcopyright{acmlicensed}
\acmConference[Woodstock '18]{Woodstock '18: ACM Symposium on Neural Gaze Detection}{June 03--05, 2018}{Woodstock, NY}
\acmBooktitle{Woodstock '18: ACM Symposium on Neural Gaze Detection, June 03--05, 2018, Woodstock, NY}
\acmPrice{15.00}
\acmDOI{10.1145/1122445.1122456}
\acmISBN{978-1-4503-9999-9/18/06}

\begin{document}

%
\title{Federated Learning for Personalized Humor Recognition}

%
\author{Xu Guo}
\affiliation{
    \institution{School of Computer Science and Engineering, Nanyang Technological University}
    \country{Singapore}
}
\email{xu008@ntu.edu.sg}
\orcid{1234-5678-9012}

\author{Han Yu}
\affiliation{
    \institution{School of Computer Science and Engineering, Nanyang Technological University}
    \country{Singapore}
}
\email{han.yu@ntu.edu.sg}

\author{Boyang Li}
\affiliation{
    \institution{School of Computer Science and Engineering, Nanyang Technological University}
    \country{Singapore}
}
\email{boyang.li@ntu.edu.sg}
\authornote{Corresponding Author}

\author{Hao Wang}
\affiliation{
    \institution{Alibaba Group}
    \country{China}
}

\author{Pengwei Xing}
\affiliation{
    \institution{School of Computer Science and Engineering, Nanyang Technological University}
    \country{Singapore}
}

\author{Siwei Feng}
\affiliation{
    \institution{School of Computer Science $\&$ Technology, Soochow University}
    \country{China}
}

\author{Zaiqing Nie}
\affiliation{
    \institution{Institute for AI Industry Research, Tsinghua University}
    \country{China}
}

\author{Chunyan Miao}
\affiliation{
    \institution{School of Computer Science and Engineering, Nanyang Technological University}
    \country{Singapore}
}
\email{ascymiao@ntu.edu.sg}
\authornote{Corresponding Author}

\renewcommand{\shortauthors}{Xu and Han, et al.}

%
\begin{abstract}
Computational understanding of humor is an important topic under creative language understanding and modeling. It can play a key role in complex human-AI interactions. The challenge here is that human perception of humorous content is highly subjective. The same joke may receive different funniness ratings from different readers. This makes it highly challenging for humor recognition models to achieve personalization in practical scenarios. Existing approaches are generally designed based on the assumption that users have a consensus on whether a given text is humorous or not. Thus, they cannot handle diverse humor preferences well. In this paper, we propose the FedHumor approach for the recognition of humorous content in a personalized manner through Federated Learning (FL). Extending a pre-trained language model, FedHumor guides the fine-tuning process by considering diverse distributions of humor preferences from individuals. It incorporates a diversity adaptation strategy into the FL paradigm to train a personalized humor recognition model. To the best of our knowledge, FedHumor is the first text-based personalized humor recognition model through federated learning. Extensive experiments demonstrate the advantage of FedHumor in recognizing humorous texts compared to nine state-of-the-art humor recognition approaches with superior capability for handling the diversity in humor labels produced by users with diverse preferences.
\end{abstract}

%
%


%

%

%
\maketitle

\section{Introduction}
\begin{figure}[t!]
	\centering
	\includegraphics[trim=0cm 4cm 0cm 4cm, clip=true,width=1\columnwidth]{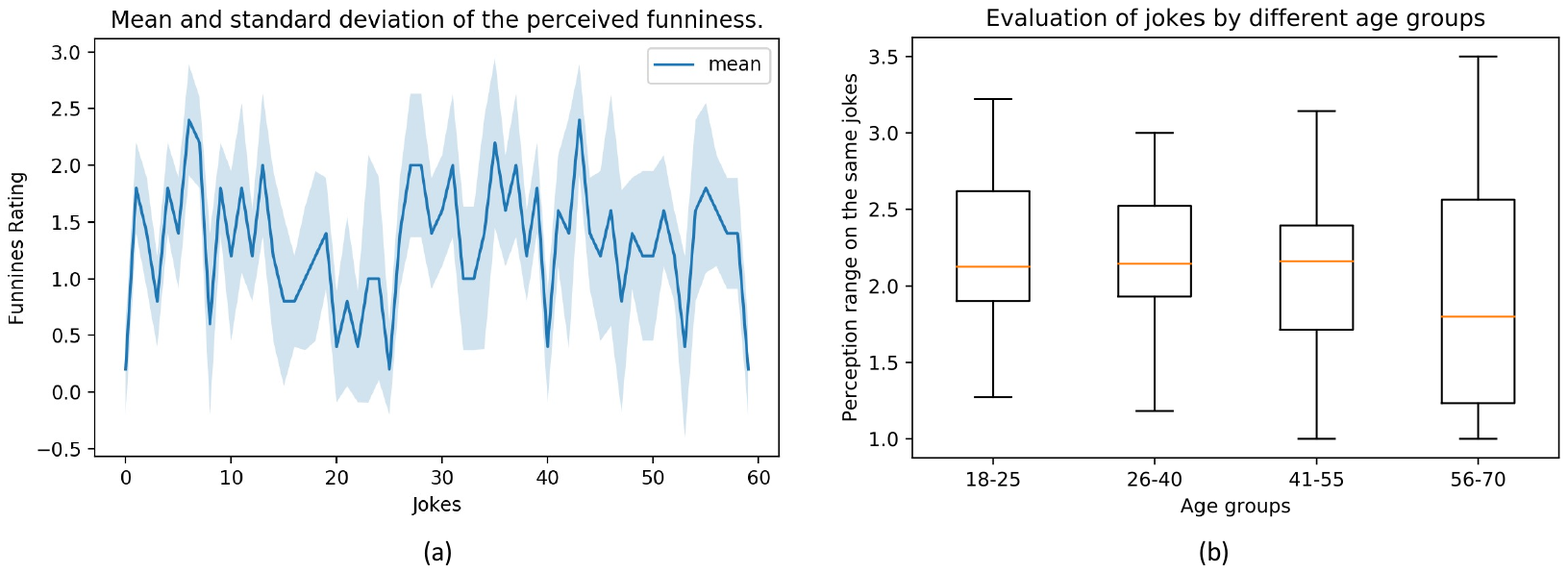}
	\caption{Empirical analysis of a random set of 60 jokes from a real-world humor rating dataset reflecting non-trivial subjectivity in human perception. (a) shows that users' perceived funniness on the same jokes vary from person to person and the variance differs from joke to joke (shaded area). (b) shows that the effect in (a) is consistent across different age groups, albeit at different levels of variance.}
	\label{empirical_analysis}
\end{figure}

In the past decade, we have witnessed broad and far-reaching success of artificial intelligence (AI) in novel application areas such as conversational characters, which provide increasingly human-like interactions to users. One particularly challenging issue in the understanding of human language is the modeling of humor. Humor is a prominent example of linguistic creativity \cite{veale2012exploding} and plays a vital role in human communication. Unlike AI tasks with objective ground truth, the task of humor recognition is characterized by the subjectivity in humor understanding. Due to individual differences in cognitive processes, the same joke can be perceived differently by its audience \cite{aykan2018assessing,martin2003individual,heintz2019four}. This is empirically proved via data analysis on a real-world dataset \cite{hossain2019president} in which each joke is rated by five persons in terms of its funniness, and the results show that the variance of human's perceived funniness on the same joke is nontrivial (Figure \ref{empirical_analysis}).

\begin{figure}[t!]
	\centering
	\includegraphics[width=1\columnwidth]{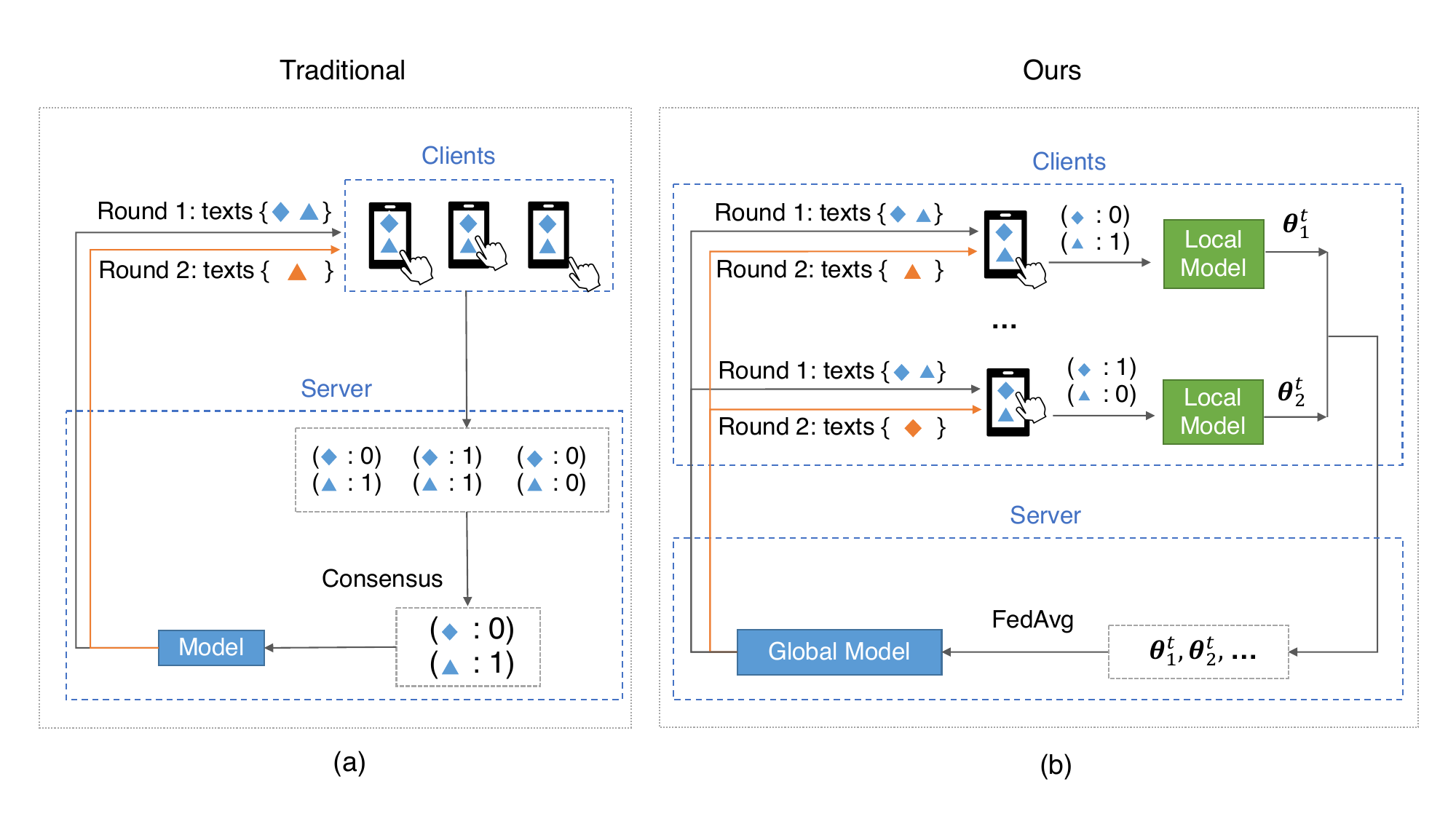}
	\caption{Traditional setting versus personalized federated learning setting for training and applying a humor recognition model (\textit{Best viewed in color}). In (a), a humor recognition model is trained on a centralized dataset whose labels are determined by majority voting (consensus) from round 1; the trained model is used to recommend funny texts to all clients at round 2, without distinction. In (b), a humor recognition model is trained on distributed datasets where the individual labels and distributions are preserved on local devices at round 1; the trained model will recommend texts to each client at round 2, possibly in different sequences.}
	\label{introduction}
\end{figure}

Computational approaches for recognizing humor generally regard the task as a binary classification problem \cite{yang2015humor,chen2018humor}. Traditional research mainly focuses on extracting expressive features as inputs to a classifier in order to improve classification performance. Neural networks such as Convolutional Neural Networks (CNNs) \cite{chen2018humor} and transformer networks \cite{mao2019bert} have shifted the focus from feature engineering to automatic feature extraction. Existing humor recognition models are generally based on the assumption that users have a consensus about whether or not a given text is humorous (as illustrated in Figure \ref{introduction}(a)) \cite{yang2015humor,liu2018exploiting,chen2018humor,zhang2014recognizing,mao2019bert}. However, in reality, humor preference is highly subjective. As a result, these approaches perform well recommending humorous contents with popular appeal, but cannot achieve personalized humor recognition. 

In this paper, we bridge this important gap in the humor recognition literature by proposing a personalized humor recognition approach - FedHumor. We adopt the assumption that humor labels from different users regarding the same text contents can be diverse, and formalize the problem as a conditional binary classification task. The federated learning (FL) architecture is used as the basis for our proposed personalized humor recognition model. 

FedHumor extends the popular Federated Averaging (FedAvg) algorithm \cite{pmlr-v54-mcmahan17a} with a diversity adaptation strategy to enhance the handling of disparate user preferences in humor recognition. As a result, it can adjust the order of humor contents (e.g., jokes) recommended to different users according to their individual preferences (as illustrated in Figure \ref{introduction}(b)). The federated training process allows multiple views over diverse label distributions, thus enhancing generalization of the humor recognition model. Meanwhile, the distributed learning process allows personalized adaptations to different users' preferences to be learned locally, thus enhancing the personalization performance of the model. Moreover, this approach does not require sensitive data concerning each user's personal humor preference to be exposed, complying with the General Data Protection Regulation (GDPR) requirement on data privacy preservation. 

To the best of our knowledge, FedHumor is the first federated learning based humor recognition model that explicitly considers the diversity of humor preferences. We conducted the personalized humor recognition by transforming existing humor recognition datasets labeled with explicit ratings into the dataset labeled with implicit binary labels reflecting different humor preferences. Extensive experiments against nine state-of-the-art approaches show that FedHumor achieves the best performance, surpassing the best performing existing approach by 2.5\% in terms of $F_1$ score. FedHumor represents a promising direction for personalized recommendation of humorous content under tightened data privacy protection regulations \cite{FL:2019}, thereby enabling innovative forms of human-AI interaction to emerge. 

Our contributions are summarized as follows:
\begin{itemize}
    \item We relaxed the common assumption in the humor recognition literature that users have a consensus about whether or not a given text is humorous. 
    \item We proposed an FL-based solution with a personalized adaptation strategy to enable personalized humor recognition with good generalization, while not exposing private humor preference data. 
    \item We conducted extensive experiments to evaluate our proposed approach. Results show that our approach is significantly superior to existing approaches in terms of personalizing humor recognition.
\end{itemize}

\section{Related Work}

\subsection{Humor Recognition}
Based on well-established computational humor theory \cite{Shultz:1976,Gruner:1997,binsted2006computational}, early research works on text-based humor recognition mainly rely on heuristic rules to manually extract features for a classifier to detect humor \cite{yang2015humor,taylor2004computationally,liu2018exploiting}. Humor-related features can be broadly divided into stylistic features and linguistic structures. Stylistic features such as alliteration, antonyms, and slang \cite{mihalcea2005making} are prevalent in stereotypical one-liners in online repositories. Linguistic structures are employed to quantitatively analyze the existence of humor patterns in a sentence \cite{taylor2004computationally,liu2018exploiting}. These patterns are exploited to reflect humor in aspects such as incongruity, ambiguity, interpersonal effects, and phonetic style \cite{yang2015humor}. Nevertheless, these manual feature engineering methods are labor-intensive. 

With the advance in computational power and the explosion of data, deep learning has been widely adopted to assist or substitute traditional feature engineering in many areas. Compared with other fields, the development of deep learning for humor recognition is more recent. The existing line of research can be divided according to the milestones of deep learning for natural language processing (NLP). The first category is to design neural architectures, such as CNNs, to learn sentence representations from pretrained word embeddings (e.g., GloVe) \cite{chen2017convolutional,chen2018humor}. Another category is to fine-tune pretrained language models (e.g., BERT) on humor recognition datasets \cite{mao2019bert}. Nevertheless, these techniques did not bring new perspectives on humor studies. 
Different from previous research, this paper is concerned with the issue of label diversity in existing humor recognition datasets, as the existing datasets are all labeled by a few annotators. For example, One Liners \cite{mihalcea2005making}, Pun of the Day \cite{yang2015humor}, The Onion News \cite{mihalcea2007characterizing}, SemEval-2017 shared task 7 \cite{miller2017semeval} and \cite{zhang2014recognizing} were collected for binary classification. The SemEval-2017 shared task 6 \cite{potash-etal-2017-semeval} is a tweets dataset annotated with three classes determined via ranking. These datasets generally overlooked the differences between individual preferences. Moreover, the non-humorous samples they choose are simply selected from other normal texts, which puts the model at a risk of learning shortcut features \cite{geirhos2020shortcut} rather than understanding the humor in texts. The recent SemEval-2020 shared task 7 \cite{hossain2019president} avoided these issues by creating a dataset in which the jokes are produced by some users and the funniness scores for those jokes are the average scores rated by other crowd-sourced users. However, this dataset is not dedicated for personalization. In this paper, we will reuse this dataset for our studies.

In psychological literature, it has been shown that the gap between the intended humor and the perceived humor is non-trivial \cite{wimer2008expectations}. The lack of label diversity in existing humor datasets precludes existing models from learning such personalized preferences. This paper attempts to address this problem by leveraging federated learning to enhance personalized humor recognition.

\subsection{Federated Learning and Personalization}
FL aims to protect data privacy by facilitating joint training of a global model across multiple data silos without exposing local data \cite{kairouz2019advances,yang2019federated}. FL can be divided into two major categories: horizontal FL and vertical FL. Horizontal FL involves data owners with local datasets in the same feature space, while vertical FL involves data owners with local datasets in different feature spaces. Our work belongs to the category of horizontal FL. The mainstream research works in horizontal FL generally assume an even partitioning on data samples and that $P(y|\bm x)$ holds the same across devices as in FedAvg \cite{pmlr-v54-mcmahan17a}, FedProx \cite{li2018federated}, FedOpt \cite{asad2020fedopt} and many other FL problems such as incentive mechanism research for fairness \cite{yu2020fairness, lyu2020towards}. Therefore, these approaches only develop common outputs for all users. In many scenarios, the assumption of independent and identically distributed (i.i.d.) samples does not hold across different data silos. 

Existing works on heterogeneous horizontal FL can be divided into two categories. The first category of research aims to eliminate statistical heterogeneity to ensure robust convergence of the global model. For example, FedProx \cite{li2018federated} added a weight approximation term to restrict the solutions of the local models to be closer to that of the global model. In \cite{chen2020focus}, a credit-weighted weights averaging approach is proposed to handle label disparities arising from healthcare practitioners' varying skill levels. The second category aims to adapt to the naturally occurring data heterogeneity under FL settings and personalize the global model for each data owner.

There are mainly two ways to achieve personalization. Firstly, we can adapt the global FL model weights for each data owner. For example, in Federated Multi-task Learning \cite{smith2017federated}, a global model is optimized on all data silos using FedAvg while each device learns a separate model with the whole model's weights \cite{li2020federated} or with only the final layers' weights \cite{arivazhagan2019federated} being optimized on the local dataset. On top of this, meta-learning has also been used to find better local solutions for each data owner \cite{fallah2020personalized}. Secondly, the model predictions or output logits can be adapted for each data owner. For example, \cite{ramaswamy2019federated} proposed a diversification mechanism to tune the diversity of the predicted emoji hints made by FedAvg during users' typing. Our work belongs to this category. 

A common default assumption made in horizontal FL is that the datasets are partially owned by each user and their features are heterogeneously distributed while the FL model is forced to learn across each user's device to build a robust machine learning model. Different from this assumption, we focus on the problem in which the same input data (i.e., humorous texts) are labeled differently by different users when viewed on their own devices.

\section{Preliminaries}
\subsection{Notations}
In this paper, we use bold letters for vectors and matrices, plain capital letters for sets and tuples, and plain lowercase letters for scalars. The probability of an event is denoted by $P$ and the probability density function of a random variable is denoted by $p$. $|\cdot|$ indicates set cardinality while $||\cdot||$ indicates the Euclidean norm of a vector or matrix. We use Greek letters in the subscript to indicate function parameters. For example, $f_{\bm \theta}$ denotes a function parameterized by $\bm \theta$. $\mathbb{E}_{p(\bm x)}[f\bm x)]$ denotes taking the expectation of $f(\bm x)$ over the distribution $p(\bm x)$. The nabla $\bigtriangledown$ indicates the gradient. For example, $\bigtriangledown_{\bm \theta}\mathcal{L}(\bm \theta)$ denotes the gradient of $\mathcal{L}(\bm \theta)$ with respect to the parameters $\bm \theta$.

\subsection{Federated Gradient-based Optimization}
\label{fedDefinition}
We find the optimal parameters $\bm \theta^{*}$ for the function $f_{\bm \theta}$ by optimizing a differentiable loss function $\mathcal{L}(\bm \theta)$ on a training set $\mathcal{D}$, where $\bm x \sim p(\bm x)$, $\forall \bm x \in \mathcal{D}$. 

\begin{equation}\label{eqn:1}
    \bm \theta^{*} = \argmin_{\bm \theta} \mathcal{L}(\mathcal{D}; \bm \theta)
\end{equation}

The optimal $\bm \theta^{*}$ is found by iterated gradient descent. At each step $t$, we update $\bm \theta$ with learning rate $\eta$ controlling the step size in the direction of the gradient $\bigtriangledown_{\bm \theta^t} \mathcal{L}(\mathcal{D};\bm \theta^t)$:

\begin{equation}\label{eqn:2}
    \bm \theta^{t+1} \leftarrow \bm \theta^{t} - \eta \bigtriangledown_{\bm \theta^t} \mathcal{L}(\mathcal{D};\bm \theta^t)
\end{equation}

\noindent The above update is repeated until convergence or a predefined number of iterations is reached. 

Under the Federated Learning setting, $m$ participants in the federation wish to jointly optimize a model without sharing data with each other. Federated Averaging (FedAvg) \cite{pmlr-v54-mcmahan17a} provides a general method to perform this optimization across $m$ data silos, $\mathcal{D}=\bigcup_{i=1}^{m}\mathcal{D}_i$ and $\bm x \sim p^{(i)}(\bm x)$, $\forall \bm x \in \mathcal{D}_i$. The local model parameters are adapted by gradient descent from global parameters $\bm \theta^{t}$ at each step. To distinguish local parameters and global parameters, we let $\bm \theta^{t, k}_i$ denote the local parameters at the $i^{\text{th}}$ participant after applying $k$ local updates to the global parameter $\bm \theta^{t}$. 
\begin{equation}
\label{eqn:3}
\begin{split}
    \bm \theta_{i}^{t, 0} & \leftarrow \bm \theta^t \\ 
    \bm \theta_{i}^{t, k+1} & \leftarrow \bm \theta_{i}^{t, k} - \eta_t \bigtriangledown_{\bm \theta_{i}^{t,k}} \mathcal{L}(\mathcal{D}_i; \bm \theta_{i}^{t, k})
\end{split}
\end{equation}

\noindent After every $K$ local iterations, we synchronize the parameters across participants by averaging local updates. The global parameter update can be written as

\begin{equation}\label{eqn:4}
    \bm \theta^{t+1} \leftarrow \frac{1}{m} \sum_{i=1}^{m} \bm \theta_{i}^{t, K}
\end{equation}

\begin{figure}[!t]
	\centering
	\includegraphics[trim=0cm 1cm 0cm 1cm, clip=true,width=0.6\columnwidth]{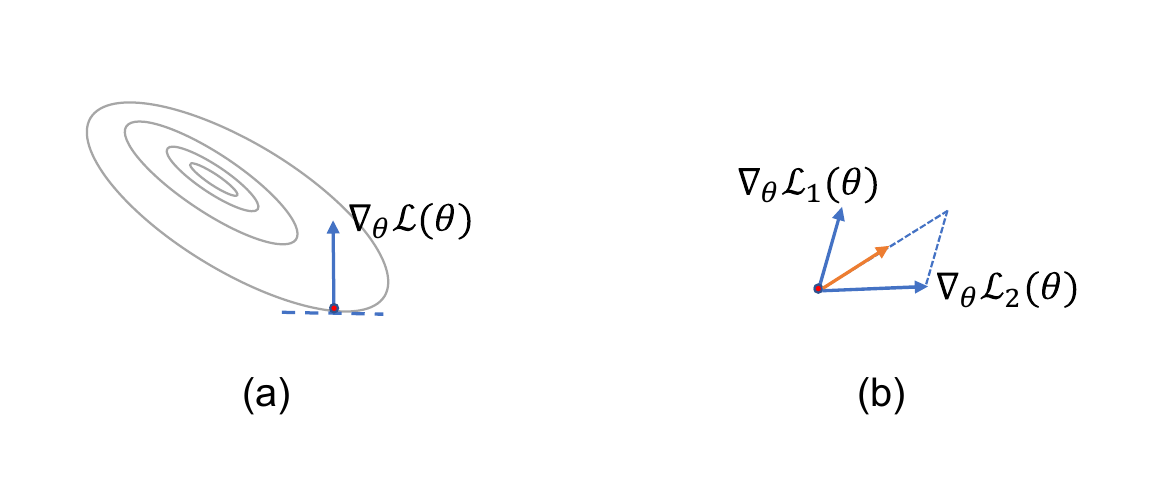}
	\caption{Model parameters updated following standard gradient descent (a), and averaged gradient descent in federated learning (b). \textit{Best viewed in color.}}
	\label{gradient}
\end{figure}

This implies that the model parameters are updated following the direction of the sum of the gradient vectors as illustrated in Figure \ref{gradient}. Traditionally, the parameters are updated iteratively on batches of labeled samples randomly sampled from the training set. In this case, the gradient descent direction is guided by the loss over a batch of observed samples (Figure \ref{gradient}(a)). In federated learning, multiple batches of samples are observed by the global model during every training iteration, while only a compromised step (i.e. $\frac{1}{2}(\bigtriangledown_{\bm \theta}\mathcal{L}_1(\bm \theta) + \bigtriangledown_{\bm \theta}\mathcal{L}_2(\bm \theta))$) is taken to update parameters $\bm \theta$ (Figure \ref{gradient}(b)).

\section{Problem Formulation}

Consider a practical scenario in which a conversational character wants to tell a user a joke and hence needs to predict whether the user will perceive the joke as humorous or not. The conversational character has a set of historical texts that have binary feedback from a set of users, who may disagree with each other on if a joke is funny. Instead of attempting to reconcile the differences of the user-generated labels, in this paper, we study a personalized humor recognition problem in which the predicted 
classes of jokes are conditioned on the personal preferences of users. 


Formally, $\bm X=(\bm x_{1}, ..., \bm x_{n})$ denotes a total of $n$ input jokes, where $\bm x_j \sim p(\bm x)$ is determined by a content publisher. $Y_i=\{y^{(i)}_{1}, ..., y^{(i)}_{n}\}$, where $y_{j}^{(i)}\in\{0, 1\}$, denotes the target labels produced by user $i$ on inputs $\bm X$. We use $\alpha$, a quantified funniness threshold lying between the liked and disliked jokes in the historical data of a user, to denote personal humor preference. For example, the distribution $p(y|\bm x, \alpha_i)$ over the binary labels of user $i$ is determined by $\alpha_i$. Given $m$ users' historical data, the task here is to predict $Y$ for a user having $\alpha \sim p(\alpha)$. We treat this task as a conditional binary classification task.

In a standard supervised learning setting, the probability of input $p(\bm x)$ is assumed to be unchanged. A traditional binary classification task tries to train a model $f_{\bm \theta}$ to estimate the true $p(y|\bm x)$ based on sufficient observations in order to construct the relationship between data and labels $p(\bm x, y)$:

\begin{equation}\label{eqn:5}
    p(\bm x, y) = p(y|\bm x)p(\bm x).
\end{equation}

\noindent Note that the standard learning task assumes that the input data $\bm x \sim p(\bm x)$ while the label $y \sim p(y|\bm x)$ only depends on the input data $\bm x$. When applied to existing humor recognition datasets, it involves an underlying assumption that different users follow the same label distribution $p(y|\bm x)$. However, this assumption may not hold for humor perception in practice. 

Here, we assume that $p(y|\bm x)$ relates to users' preference $\alpha$. The problem thus becomes training the model $f_{\bm \theta}$ to estimate the conditional probability $p(y | \bm x, \alpha)$ based on given $\alpha \sim p(\alpha)$ and observed $\bm x \sim p(\bm x)$:

\begin{equation}\label{eqn:6}
    p(\bm x, y, \alpha) = p(y | \bm x, \alpha) p(\bm x,\alpha) = p(y | \bm x, \alpha) p(\bm x) p(\alpha).
\end{equation}

\noindent Here, $p(\bm x)$ is determined by the content publisher and $p(\alpha)$ is determined by users. So they are independent and satisfy: $p(\bm x,\alpha)=p(\bm x)p(\alpha)$.

\section{The Proposed Model} \label{FedHumor}
In this section, we describe the detailed design of the proposed FedHumor model for the task of personalized humor recognition. We first introduce the contextualized text encoder model, then we present the federated learning-based model training, followed by the diversity adaptation design for different users, and finally we introduce the model validation under FL setting. The pseudo-code for FedHumor is given in Algorithm \ref{alg:algorithm}. 

\subsection{FedHumor Network}
We employed the pretrained language model - BERT \cite{devlin-etal-2019-bert} - to capture contextualized sentence representations as the input features to a classification layer, and fine-tune the pretrained weights together with classifier parameters on our task. Briefly, BERT tokenizes a piece of text into a sequence of word IDs indexed by its stationary vocabulary. The IDs are used to fetch their corresponding word embeddings from an embedding table: $\bm W=(\bm W_1,...,\bm W_l)$, $\bm W \in \mathbb{R}^{l\times d}$. $l$ is the fixed maximum length of a sentence and $d$ is the word embedding size. The interactions between words and order information are captured through the self-attention modules in twelve transformer layers, denoted as $\bm T$. The first vector ([CLS]) of the final hidden states at the last transformer layer is used to represent the final contextualized sentence representation, $\bm x$, for the given sentence. A non-linear classification layer, denoted as $\bm C$, followed by a $tanh$ activation function, accepts the sentence representations and makes predictions. The trainable parameters in the FedHumor model are $\bm \theta=[\bm W;\bm T;\bm C]$ and time complexity for each round is $\mathcal{O}(|\bm W||\bm T||\bm C|K)$. $K$ is the local training iterations on each user's device while all users perform local adaptation synchronously.

\subsection{Weight-tying Federated Training} \label{FedTrain}
\begin{figure}[!t]
	\centering
	\includegraphics[width=0.7\columnwidth]{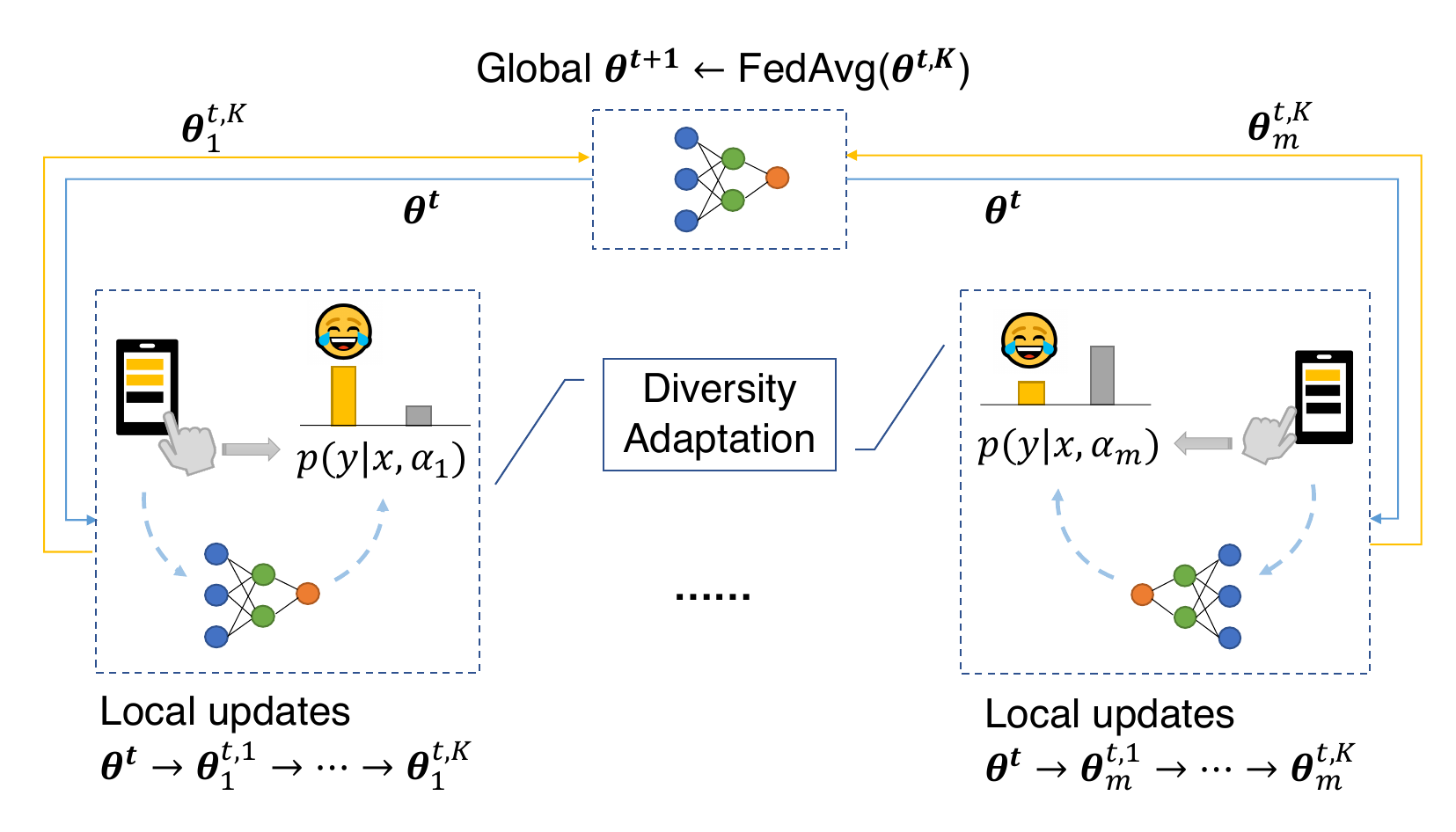}
	\caption{The training of FedHumor involves three steps: 1) server sends the global model to clients; 2) the clients train the model locally based on their own labels, and send their updated parameters to server; 3) server aggregates local updates to produce a new global model. }
	\label{FL_train}
\end{figure}

We follow the FedAvg algorithm \cite{pmlr-v54-mcmahan17a} to aggregate the model parameters updated on each user's data silo $\mathcal{D}_i=\{\bm X, Y_i\}$ after each round of federated training to produce a global model, as illustrated in Figure \ref{FL_train}. In every training round, the current global FL model with parameters $\bm \theta$ is sent to $m$ users randomly selected from the device population, and the adapted local parameters are denoted as $\bm \theta_i$. For each user whose humor preference is $\alpha_i$, we use cross entropy as our loss function, and the instance loss for user $i$ is computed as follows:

\begin{equation}\label{eqn:7}
\ell(\bm \theta_{i};\alpha_i)=-y\mathrm{log}(\Tilde{p}(y|\bm x, \alpha_i)) - (1-y)\mathrm{log}(1-\Tilde{p}(y|\bm x, \alpha_i)) + \lambda||\bm \theta_{i}||_2^{2}.
\end{equation}

\noindent Here, $\Tilde{p}(y|\bm x, \alpha_i))$ is the adapted estimated probability computed using Equations \eqref{eqn:10} and \eqref{eqn:11}. $\lambda$ is the hyperparameter controlling L2 norm regularization over local model parameters $\bm \theta_{i}$ (a.k.a. weight decay). $\bm \theta_{i}$ is determined by gradient descent to minimize the following training loss on user $i$'s data silo:

\begin{equation}\label{eqn:8}
\mathcal{L}(\bm \theta_{i}; \alpha_{i}) = \frac{1}{|\mathcal{D}_i|}\sum_{\mathcal{D}_i}\ell(\bm \theta_{i};\alpha_i)
\end{equation}

In each federated training round $t$, the local model parameters are adapted from $\bm \theta^t$ to $\bm \theta_{i}^{t,K}$ for $K$ local iterations. The global model parameters are adapted from $\bm \theta^t$ to $\bm \theta^{t+1}$ by averaging the aggregated parameters:

\begin{equation}\label{eqn:9}
    \bm \theta^{t+1} \leftarrow \bm \theta^{t} - \eta_t \cdot \frac{1}{m}\sum_{i=1}^{m} \bigtriangledown_{\bm \theta_{i}^{t,K}}\mathcal{L}(\bm \theta_{i}^{t,K}; \alpha_{i})
\end{equation}

\noindent where $m$ is the number of users joining the federated learning and $\eta_t$ is the learning rate at each round and its value depends on the learning rate scheduler. Throughout the process, users' personal humor preference information (i.e. $p(y|\bm x, \alpha)$) are kept on their own devices.

\subsection{Diversity Adaptation} \label{diversity}
The purpose of applying federated learning to humor recognition is to enhance the generalizability of traditional humor recognition models when dealing with diverse user preferences while preserving data privacy. FedHumor is designed similar to Google's GBoard scenario \cite{ramaswamy2019federated} based on the FL paradigm. Inspired by the mechanism employed in GBoard to prevent the federated model being dominated by highly frequent emojis, we adapt the model predictions, $\widehat{p}(y|\bm x)$, on each class $c=1,..,C$, for user $i$ as follows:

\begin{equation}\label{eqn:10}
 v_c = \frac{\widehat{P}(y=c|\bm x)}{P(y=c|\bm x, \alpha \in [\alpha_i, \alpha_i + \tau))^{\beta_i}}
\end{equation}

\noindent where $P(y|\bm x,\alpha \in [\alpha_i, \alpha_i + \tau))$ is the empirical marginal label distribution from users whose threshold $\alpha$ falls in the range $[\alpha_i, \alpha_i + \tau)$. $\tau$ is a small real number interval that reflects a significant change in label distribution when the preference varies. For example, $\forall \alpha \in [1.0, 1.1)$ is deemed the same preference, and differs from  $\forall \alpha \in [1.1, 1.2)$. For brevity, we use $\alpha=\alpha_i$ to denote an established humor preference. $\beta_i$ is a user-specific scaling factor tuned on the validation set. Although this introduces one tunable hyperparameter for every user, the sensitivity analysis in Section \ref{sensitivity} shows that it is necessary to tune $\beta_i$ only when $\alpha_i$ takes on extreme values, which result in very imbalanced label distributions. The new probability is computed using the softmax function: 

\begin{equation}\label{eqn:11}
    \Tilde{P}(y=c|\bm x, \alpha=\alpha_i) = \frac{\exp({v_c})}{\sum_{j=1}^{C} \exp({v_{j}})}.
\end{equation}

Intuitively, this approach can penalize predictions that are too confident about a certain class due to its dominant frequency in the observed samples. It is based on the heuristic that users may possess different levels of arousal in response to the same humorous content. Those having \textit{easy-to-amuse} or \textit{hard-to-amuse} personalities can result in a drastically unbalanced dataset. For example, given $\alpha=\alpha_i$, if the predicted probability is $\widehat{P}(y=1|\bm x)=0.9$ while the empirical probability is $P(y=1|\bm x)=0.7$, the denominator serves as a punishment to adjust the prediction to be $\Tilde{P}(y=1|\bm x)\approx0.72$ for $\beta=1$ (computed using Equations \eqref{eqn:10} and \eqref{eqn:11}), which is closer to the real probability.

\subsection{Federated Model Selection}
After completing each round of FL training, the global FL model will be tested on the validation set on each user's device, $\mathcal{D}^{val}=\{\bm X^{val}, Y^{val}|\alpha\}$. In this stage, $\widehat{p}(y^{val}|\bm x^{val}, \alpha)$ is the estimated probability distribution over the implicit labels in the validation set by the model. The overall validation loss for each user is calculated as follows: 
\begin{equation}\label{eqn:12}
\mathcal{L}(\bm \theta;\alpha_i)=-\frac{1}{|\mathcal{D}^{val}|}\sum_{(\bm x, y)\in \mathcal{D}^{val}}[y\mathrm{log}(\widehat{p}(y|\bm x, \alpha_i)) + (1-y)\mathrm{log}(1-\widehat{p}(y|\bm x, \alpha_i))].
\end{equation}

\noindent The validation performance across all users is calculated as follows:
\begin{equation}\label{eqn:13}
    \mathcal{L}(\bm \theta) = \frac{1}{m}\sum_{i=1}^{m}\mathcal{L}(\bm \theta; \alpha=\alpha_i)
\end{equation}

\noindent The best global model $\bm \theta$ is selected based on the lowest validation loss, $\mathcal{L}(\bm \theta)$, among all federated rounds.

\begin{algorithm}[!t]
\vspace{6pt}
Initialize global model parameters $\bm \theta$;

\For{ $t=1,2,...,T$ }{
    \For{ each user $i=1,2,...,m$ \rm \textbf{in parallel}}{
        $\bm \theta_{i}^{t+1} \gets \rm ClientUpdate$ $(i, \bm \theta^t, \beta_i)$;
    }
    $\bm \theta^{t+1} \gets \frac{1}{m}\sum_{i=1}^{m}\bm \theta_{i}^{t+1}$: global parameters update (Eq. \eqref{eqn:9});
    
    \For{ each user $i=1,2,...,m$ \rm \textbf{in parallel}}{
         $ \mathcal{L}(\bm \theta^{t+1};\alpha_i, t) \gets \rm Inference (\textit{i}, \bm \theta^{t+1})$;
    }
    
   $\mathcal{L}(\bm \theta^{t+1}, t)=\frac{1}{m}\sum_{i=1}^{m}\mathcal{L}(\bm \theta^{t+1};\alpha_i, t)$ (Eq. \eqref{eqn:13});
}

Output: $\bm \theta^{*} \gets \rm min(\mathcal{L}(\bm \theta^{t+1}, t))$;

\vspace{6pt}

\Fn{(i,$\bm \theta$,$\beta$) $\rm:$ }{
    Prepare $D_{train}=\{\bm X, Y_i|\alpha_i\}$;
    
    \For{ $j=1,2,...,E$ }{
    
        \For{ $\mathcal{B}$ in $D_{train}$ }{
            $\widehat{p} = f_{\bm \theta}(\bm x; \alpha_i)$: make predictions;
            
            $\widetilde{p} = \mathrm{softmax}(\widehat{p} / p(y|\alpha_i)^{\beta_{i}})$: adapt predictions (Eq. \eqref{eqn:10}, \eqref{eqn:11});
            
            $\mathcal{L}(\bm \theta;\alpha_i) = \frac{1}{|\mathcal{B}|} \sum_{\mathcal{B}} \ell(\widetilde{p}, y; \bm \theta)$: compute training loss (Eq. \eqref{eqn:8});
            
            $\bm \theta_{i} \gets \bm \theta - \eta_t \bigtriangledown_{\bm \theta} \mathcal{L}(\bm \theta;\alpha_i)$: local parameters update;

        }
    }
    return $\bm \theta_{i}$
}

\vspace{6pt}

\Inference{($i,\bm \theta$)}{
\hspace{5pt}

    Prepare $D_{val}=\{\bm X, Y_i|\alpha_i\}$;

     $\widehat{p} = f_{\bm \theta}(\bm x; \alpha_i)$: make predictions;

    $\mathcal{L}(\bm \theta;\alpha_i) = \frac{1}{|\mathcal{D}_{val}|} \sum_{\mathcal{D}_{val}} \ell(\widehat{p}, y;\bm \theta)$: validation loss (Eq. \eqref{eqn:12});

return $\mathcal{L}(\bm \theta;\alpha_i)$

}

\caption{\textnormal{FedHumor}. The $m$ users are indexed by $i$; $D_{train}$ denotes training set and $D_{val}$ denotes validation set; $\alpha_i$ is the humor preference of user $i$; $\beta_i$ indicates diversity adaptation for user $i$; $\mathcal{B}$ denotes a mini-batch of $D_{train}$; $E$ is local training epochs and $T$ is total federated training rounds; $\eta_t$ is the learning rate at each round $t$.}
\label{alg:algorithm}
\end{algorithm}

\section{Experiment Setup}
In this section, we present the details of our experimental setting. We first introduce a real-world humor recognition dataset and the preparation for personalized humor recognition. Then, we introduce the evaluation metrics used for performance comparison. Finally, we describe our model setting.

\subsection{Dataset Description}
In our experiments, we use a newly published dataset from SemEval-2020 shared Task 7\footnote{SemEval 2020 Task 7 (Sub-task 1) dataset downloaded from: \url{https://github.com/n-hossain/semeval-2020-task-7-humicroedit}} - assessing the funniness of edited news headlines \cite{hossain2019president}, which contains humorous headlines created based on the incongruity theory. Each created headline was sent to five crowdsourced annotators through Amazon Mechanical Turk and each annotator was required to rate its funniness using a score from the integer interval [0, 1, 2, 3]. For example, a news headline \textit{``Royal wedding: Meghan's dress in detail''} was micro-edited by replacing $dress$ with $elbow$ to produce a funny version \textit{``Royal wedding: Meghan's elbow in detail''}, which received five ratings: 0, 1, 3, 3 and 3. The original dataset made a consensus assumption on the perceived funniness of the edited headline. They reassigned an average of the five ratings, $2.0$, to be the funniness rating for this joke. This example shows a common assumption of an average user preference does not reflect the diversity among different users. The statistics of the dataset are summarized in Table \ref{dataset}. 

\begin{table}[ht]
\caption{Statistics of the public dataset}\smallskip
\label{dataset}
\centering
\smallskip\begin{tabular}{@{}rccc@{}}
\toprule
                  & Train & Validation   & Test  \\ \midrule
Number of samples & 9,652 & 2,419 & 3,024 \\ 
Average Rating   & 0.936 & 0.935 & 0.940 \\
Minimum Rating    & 0.000 & 0.000 & 0.000 \\
Maximum Rating    & 3.000 & 3.000 & 2.800 \\\bottomrule
\end{tabular}
\end{table}

\subsection{Implicit Label Generation}\label{sec:implicit}
\begin{figure}[t]
	\centering
	\includegraphics[trim=1cm 2cm -1cm 2cm, clip=true, width=1.1\columnwidth]{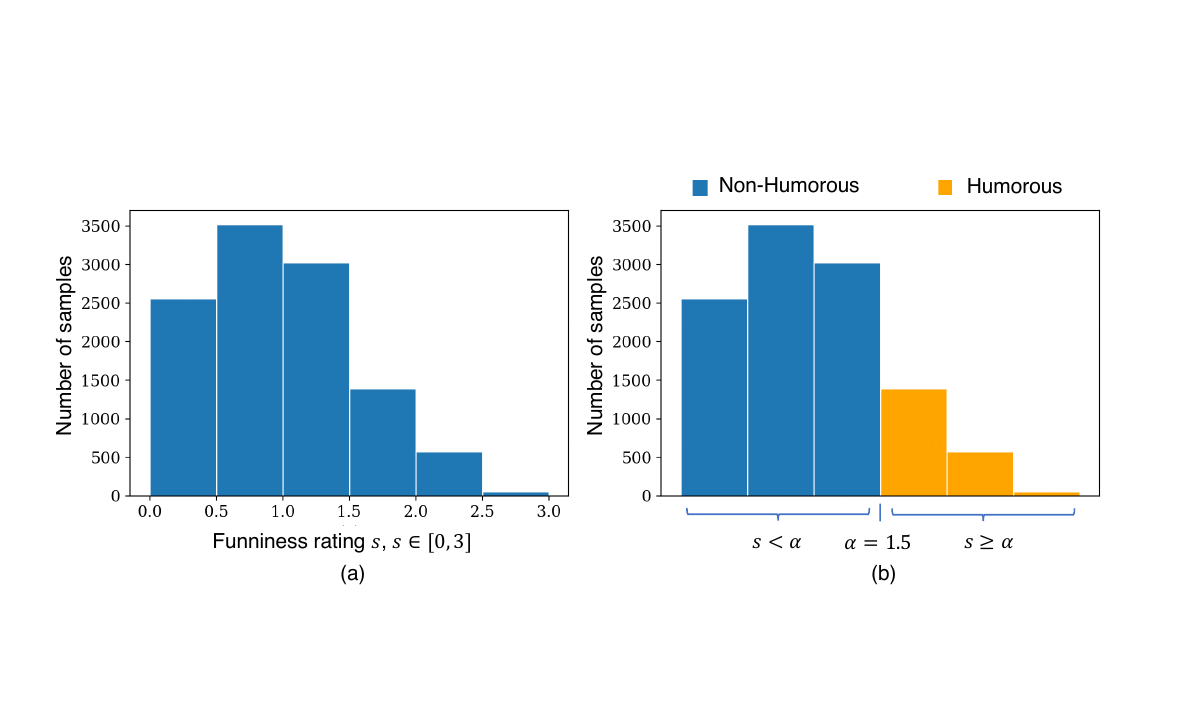}
	\caption{Transform explicit ratings into binary labels. The distribution of explicit funniness ratings on a set of jokes rated by the content publisher is shown in discrete intervals (a). An example in which a user's humor preference is quantified by $\alpha=1.5$ (b). \textit{Best viewed in color.}} 
	\label{alpha_effects}
\end{figure}

Jokes can be perceived differently and thus, receive different funniness ratings from users. Unfortunately, there are no public datasets for training a NLP model under the problem of label distribution shift across people. It is also difficult to collect true feedback from a diverse population in a privacy protection setting. To simulate the diversity in the human perception of jokes, we generate a synthetic binary humor recognition dataset to study this problem from the SemEval ratings dataset. 
\begin{enumerate}
    \item First, sort the jokes by their original average funniness ratings. Specifically, given a set of $n$ jokes rated by the content publisher with funniness ratings drawn from a non-negative ordinal interval $I=[s_{min},s_{max}]$, e.g., $I=[0,3]$. $\mathcal{D}=\{(\bm x_{j}, s_{j}))\}_{j=1}^{n}$, where $s_{j}\in I$ is the original dataset with ratings sorted in ascending order (Figure \ref{alpha_effects}(a)). Users receiving this dataset would agree on the jokes with funniness ratings that are close to the boundaries of the rating interval (i.e. not funny ($s_{min}$) and very funny ($s_{max}$)), and may disagree on the jokes with ratings lying in between (i.e. slightly funny and moderately funny ($s_{min} < s_j < s_{max}$)). 
    \item Second, assume that every user can have only one unique humor preference, denoted as $\alpha_i$. Based on a user's historical implicit feedback, the user's humor preference, $\alpha_i$, is reflected as a specific threshold within the funniness rating interval, i.e., $s_{min} < \alpha_i < s_{max}$. The user dislikes jokes below this threshold rating, $s_{j} < \alpha_i$, and likes jokes above it, $s_{j}\geq \alpha_i$ (Figure \ref{alpha_effects}(b)). In effect, due to the limited number of training samples observed for every funniness rating, $\forall \alpha_i \in [\alpha_i, \tau)$ is treated as the same humor preference when $\tau$ is small enough.
\end{enumerate}
Based on the first assumption, we create a set of implicit labels, $Y_i=\{y^{j}\}_{j=1}^{n}$ for users with $\alpha_i$ on the same jokes $\bm X$. Based on the second assumption, we can create multiple sets of implicit binary labels of diverse distributions $p(y|\bm x,\alpha_i)$, where $y_{j}=0$ for $s_{j}<\alpha_i$ and $y_{j}=1$ for $s_{j}\geq \alpha_i$.

\begin{figure}[!t]
	\centering
	\includegraphics[trim=1cm 4cm -1cm 4cm, clip=true,width=0.7\columnwidth]{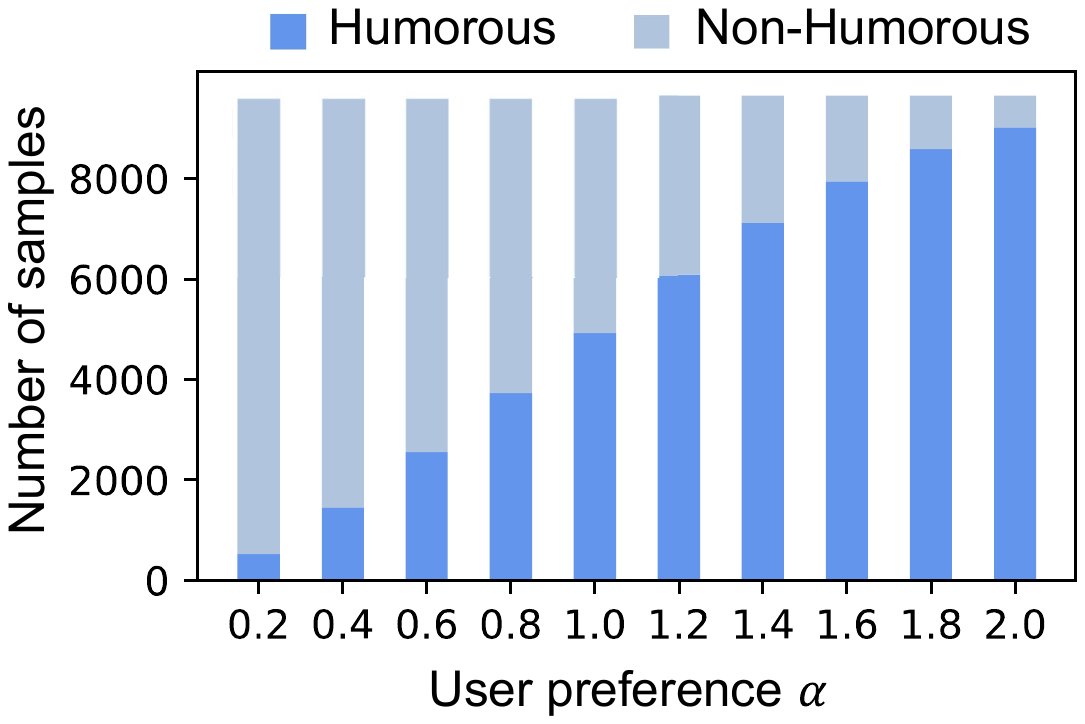}
	\caption{Generated binary labels with different distributions from funniness ratings when diverse humor preferences are considered. \textit{Best viewed in color.}}
	\label{alpha_distribution}
\end{figure}

To simulate the real-world diversity of perceived funniness, we generate a diverse population with humor preferences $\alpha$ ranging from $\alpha=0.2$ to $\alpha=2.0$ with a step size of $0.1$ (i.e., $\tau=0.1$), as illustrated in Figure \ref{alpha_distribution}. The higher the $\alpha$ value, the more non-humorous labels will be generated. In our experiments, a valid user is regarded as having $\alpha_i \in [0.2, 2.0]$. Specifically, the funniness ratings of the dataset are transformed into implicit binary labels generated by a given user with the preference $\alpha$ on his own device. The distribution over the generated implicit labels is $p(y|\textit{\textbf{x}},\alpha)$. The boundary values (e.g., $\alpha=0$ and $\alpha=3$) that result in a one-class dataset are not considered in our experiments.

\subsection{Evaluation Metrics}
To evaluate the classification performance of FedHumor and baseline approaches, we use the \textit{macro-averaged} $F_1$ score as our main metric, which is the average of the $F_1$ scores of two classes. We additionally report the \textit{macro-averaged} precision and recall scores. Note that a \textit{macro-averaged} metric implies that the generalization performance on all classes are equally important. This is essential in personalized humor recognition, because some users with very low or very high threshold of humor preference may produce very imbalanced labels. In this case, a model can achieve an artificially high accuracy by predicting all the samples to be the dominant class. Thus we do not use the accuracy metric for evaluation. All the methods are compared on the hold-out test set and their predicted implicit labels for each class are counted to calculate the \textit{macro-averaged} metrics.

\subsection{Model Setting}
FedHumor is built based on top of the base version of BERT \cite{devlin-etal-2019-bert}. The pretrained weights we utilize include its word embedding layer and twelve transformer layers\footnote{PyTorch implementation of the pretrained BERT model : \url{https://huggingface.co/transformers/pretrained_models.html}}. A pooling layer nonlinearized by $tanh$ activation function\footnote{PyTorch implementation of the tanh activation function: \url{https://pytorch.org/docs/stable/generated/torch.nn.Tanh.html}} is applied on the first ([CLS]) token representation from the last transformer layer to be the final contextualized sentence representation. A dropout of 0.1 is applied on the pooled output and is further sent to a classifier, which is a fully connected layer randomly initialized from a uniform distribution, $\mathcal{U}(-\sqrt{1/d}, \sqrt{1/d})$, where $d$ is the size of the input features\footnote{PyTorch initialization of linear layer:  \url{https://pytorch.org/docs/stable/generated/torch.nn.Linear.html}}. In our model, $d=768$, the length of the hidden vector corresponding to the $\mathrm{[CLS]}$ token. We follow the recommended hyperparameter setting introduced in the paper \cite{devlin-etal-2019-bert} to tune learning rate, batch size and weight decay for fine-tuning BERT on downstream tasks based on Federated Averaging setting\footnote{Code downloaded from: \url{https://github.com/shaoxiongji/federated-learning}}.

\section{Performance Evaluation}
In this section, we conduct experiments to evaluate the performance of FedHumor. We compare it against nine state-of-the-art humor recognition methods on the hold-out test set. We also study the properties of FedHumor through experiments and compare it with other learning approaches to show the advantages of following the federated learning paradigm. 

\subsection{Comparison of Different Training Strategies}
Given that the property of our data is Differently and independently distributed, we first compare different training strategies against the federated weight-tying method.  
\begin{enumerate}
    \item \textbf{INDV}: A simple approach to this problem is to fine-tune a BERT model for each \textit{individual} user based on their local ground truth labels. This method represents the traditional way of providing personalized humor recognition, which creates a unique model for each user. Yet, it’s computationally expensive and might easily overfit to small imbalanced datasets in practice. Moreover, when new users registered, server has to establish new model to be trained for them.
    \item \textbf{AGG}: The second approach is to first \textit{aggregate} all the labeled data from each user into a central database, and treat all of them as ground truth labels to train a BERT-based humor recognition model. This follows the traditional way of training a single-task machine learning model. However, this approach needs to have access to user’s data, which may cause privacy breach.
    \item \textbf{FED}: he third approach is to tie the weights of each individual BERT model through \textit{federated learning}. In this setting, users can share learned BERT parameters while keeping their personal adaptation module. 
\end{enumerate}

To test these approaches under different levels of preference diversity, we prepare two groups of user preferences as shown below. Due to the limit of dataset size and the funniness score range, the maximum number of different user preferences we can generate from the given dataset is 18.

\begin{itemize}
    \item Group 1: a group of three users with different preferences: $\alpha=0.3$ represents the easy-to-amuse personality, $\alpha=0.9$ represents the neutral personality, and $\alpha=1.8$ represents the hard-to-amuse personality; 
    \item Group 2: a group of 18 users with different preferences: $\alpha$ ranging from $0.2$ to $1.9$ in increments of $0.1$, representing a more diverse population.
\end{itemize}

We compare the average test performance of a group between the three models. Comparison Results are shown in Table \ref{Learning_strategy_comparison}. Take the column of F score as an example. Aggregated training approach works the worst, meaning that training data with conflict labels can degrade the performance. Individual training comes after, meaning that such issue can be alleviated by separating them. Federated training further improves the performance, meaning that allowing weight-tying between different users in the federated setting can bring performance gains. Comparing the two groups, the more diverse the users, the better the Federated Learning approach performs compared to the rest.
 
\begin{table}[h]
\caption{Average test performance (in \%) of three learning strategies on two groups of users. Values in bold denote the best results. Underlined values indicate the second best results.}\smallskip
\label{Learning_strategy_comparison}
\centering
\begin{tabular}{@{}cllll@{}}
\toprule
\multicolumn{2}{l}{} & Precision & Recall & $F_1$ score \\ \midrule
\multirow{3}{*}{Group 1} & AGG & \underline{58.59} & 54.89 & 41.66 \\
 & INDV & 56.30 & \underline{55.32} & \underline{53.52} \\
 & FED & \textbf{60.03} & \textbf{65.57} & \textbf{55.61} \\ \midrule
\multirow{3}{*}{Group 2} & AGG & 57.40 & 51.25 & 33.05 \\
 & INDV & \underline{58.14} & \underline{55.61} & \underline{53.03} \\
 & FED & \textbf{61.67} & \textbf{66.62} & \textbf{57.48} \\ 
 \bottomrule
\end{tabular}
\end{table}

\subsection{Hyperparameter Sensitivity Analysis} \label{sensitivity}
As introduced in Section \ref{diversity}, we allow a hyperparameter $\beta$ on each user to control how much we want to scale the model's estimated probabilities $P(y|x, \alpha_i)$ w.r.t. the marginal local distribution $P(y|x, \alpha_i)$. As such, we conduct a sensitivity analysis through grid search to reveal whether it affects the performance of FedHumor and how to set $\beta$ for each user. To do so, we vary the values of $\alpha$ from low ($\alpha=0.2$) to high ($\alpha=2.0$) and at the same time, increase the scale factor $\beta$ from small ($\beta=0$) to large ($\beta=2.0$), all in increments of $0.1$. This results in 399 experiments in total. The implicit binary labels are generated each time the $\alpha$ value changes. We train the BERT-based classification model for each combination of $\alpha$ and $\beta$ and the best corresponding model is selected on the validation set. We report the macro-averaged $F_1$ score on the hold-out test set. 
\begin{figure}[t]
	\centering
	\includegraphics[trim=0cm 4cm 0cm 4cm, clip=true, width=0.7\columnwidth]{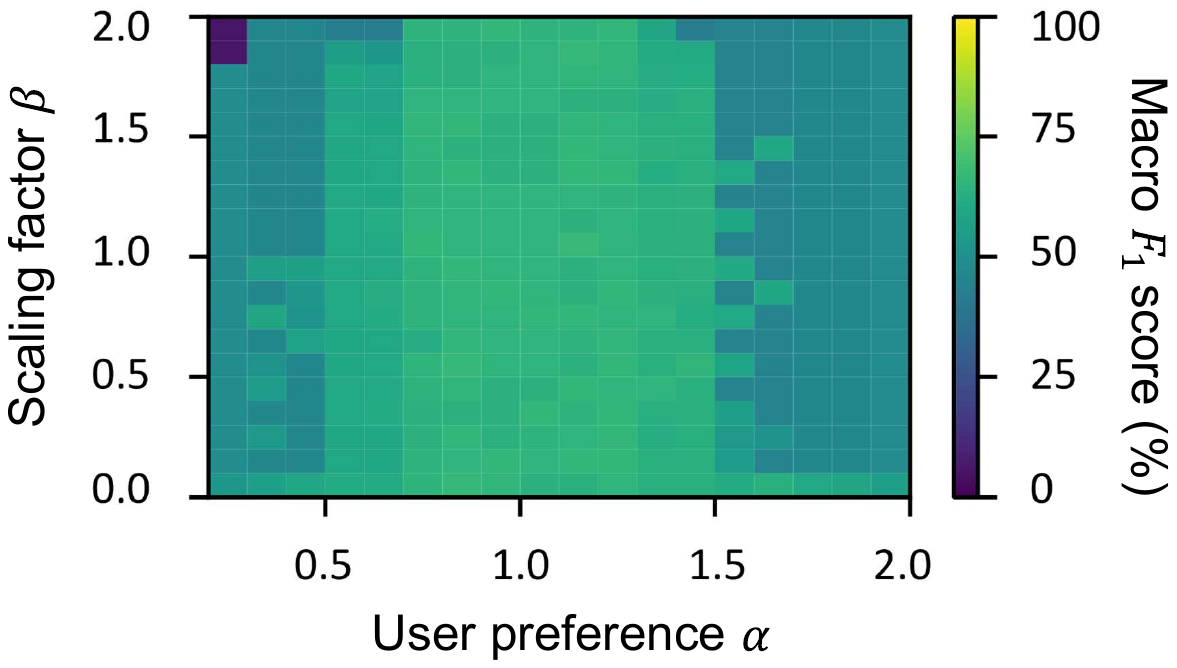}
	\caption{Tune hyper-parameter $\beta$ w.r.t. $\alpha$ for single user. \textit{Best viewed in color.}}
	\label{alpha_beta_effects}
\end{figure}

The results on the test set are shown in Figure \ref{alpha_beta_effects}. In general, a model can achieve better generalization performance on the preference range of $[0.5, 1.5]$. From Figure \ref{alpha_distribution}, we can see that this preference interval leads to the range of empirical probability from $P(y=1|\textit{\textbf{x}}, \alpha=0.5)\approx0.2$ to $P(y=1|\textit{\textbf{x}}, \alpha=1.5)\approx0.75$. In this case, for every $\alpha$ value, there are multiple $\beta$ values that can achieve the best performance. This means that our model is not very sensitive to $\beta$ and a reasonable value would suffice. For users who have very low ($\alpha<0.5$) or very high ($\alpha>1.5$) humor preference values, their labels are too imbalanced for the model to achieve good generalization performance. 
In this case, $\beta$ is recommended to set to a low value (e.g., around $0.1$).

\subsection{Comparison of Different Humor Recognition Models}
Finally, we compare our model with three kinds of baselines. The first three models are the representative humor recognition models in literature, which do not use pretrained language model. The second six models are different versions of pretrained language models with fine-tuning. Our model use both a pretrained language model and federated training on diverse humor preferences. 
\begin{enumerate}
    \item \textbf{DV-LR}: We trained Doc2Vec using the distributed bag of words approach \cite{mikolov2013distributed} for sentence representation and applied logistic regression classifier on the features for classification.
    \item \textbf{WV-RF}: We reproduced the humor recognition model in \cite{yang2015humor}, which used a pretrained word2vec\footnote{Google word2vec downloaded from: \url{https://code.google.com/p/word2vec/}} model for sentence representation and a Random Forest classifier for classification.
    \item \textbf{WV-CNN-HN}: We reproduced the deep learning-based humor recognition model in \cite{chen2018humor}, which augmented a CNN with a Highway Network for end-to-end humor recognition\footnote{PyTorch implementation of Highway Network: \url{https://gist.github.com/dpressel/3b4780bafcef14377085544f44183353}}.
    \item \textbf{BERT-FZ}: We use the BERT model pretrained on lower-cased English text with 768 hidden size and 12 transformer layers. We \textit{freeze} the pretrained model parameters and only update the parameters of a fully connected classification layer. 
    \item \textbf{BERT-FT}: Different from the previous method, we \textit{fine-tuned} the pretrained parameters together with the classifier parameters. This is a strong baseline as many downstream NLP tasks have shown improved performance through fine-tuning BERT on the task-specific datasets \cite{devlin-etal-2019-bert}. It is also the same base model adopted by FedHumor.
    \item \textbf{BERT-L/C/M}: Advanced by huge model capacity trained on large corpora, pretrained language models can provide generally effective representations across domains. As such, we adopt other versions of BERT. BERT-L is pretrained with a much \textit{larger} hidden size (1024) and deeper (24) layers. BERT-C means the true \textit{case} and accent markers are preserved when training BERT. BERT-M means a \textit{multilingual} BERT model that was pretrained on lower-cased text in the top 102 languages with the largest Wikipedia. The three models are all \textit{fine-tuned} together with the classifier to form three baseline approaches.
    \item \textbf{ALBERT}: There are also many strong pretrained language models after BERT. We adopt ALBERT \cite{lan2019albert}, which had been evaluated to have better scalability on downstream tasks than BERT. We fine-tune its pretrained parameters together with a classification layer on the humor recognition task for comparison.
\end{enumerate}

Test Results are shown in Table \ref{humor_recognition_results}. The metrics are the macro average values over all classes. Generally, pretrained language models with domain adaptive fine-tuning outperform all traditional models. Our model trained based on the relaxed assumption, exceeds all baselines across metrics. 

\begin{table}[!t]
\caption{Test performance (in \%) on the user with $\alpha=1.0$ achieved by FedHumor and all baseline approaches. Values in bold indicate the best results. Underlined values indicate the second best results.}\smallskip
	\label{humor_recognition_results}
	\centering
	\smallskip\begin{tabular}{@{}lllll@{}}
			\toprule
			& Precision & Recall & $F_1$ score \\ \midrule
			DV-LR & 53.69 & 53.67 & 53.64 \\
			WV-RF & 56.70 & 56.10 & 55.20 \\ 
			WV-CNN-HN & 56.20 & 54.70 & 51.90 \\ \midrule
			BERT-FZ & 54.15 & 53.71 & 52.53 \\ 
			BERT-FT & \underline{64.91} & \underline{64.88} & \underline{64.87} \\ 
			BERT-L & 64.48 & 64.48 & 64.47 \\ 
			BERT-C & 62.69 & 62.65 & 62.62 \\ 
			BERT-M & 62.11 & 62.08 & 62.06 \\ 
			ALBERT & 61.06 & 61.05 & 61.04 \\ \midrule
			FedHumor & \textbf{66.60} & \textbf{66.56} & \textbf{66.53} \\ \bottomrule
		\end{tabular}
\end{table}

The first three models (i.e., DV-LR, WV-RF, WV-CNN-HN) use static word representations which were widely adopted in humor recognition tasks before the advent of large pretrained language models. Their performance are generally not good on this dataset compared to pretrained language model-based methods. The following six models (i.e., BERT-FZ, BERT-FT, BERT-L/C/M, ALBERT) are BERT variants. The results achieved by BERT-FZ are much worse than the rest. This shows that the pretrained language models, whose learning phase does not take advantage of labeled data, can only provide general representations. They should be fine-tuned towards a particular task by further learning from domain-specific label information. BERT-L is close to but lower than BERT-FZ showing that larger model parameters do not bring performance gain. BERT-C and BERT-M are lower than BERT-FZ indicating the preservation of the true case and knowledge from other languages did not benefit this task. ALBERT, though tested to be more scalable to downstream tasks than BERT on some datasets, also failed to outperform BERT on this task. FedHumor achieved the best results across all the three evaluation metrics and 2.5\% relative improvement on $F_1$ score than the second best result. This reveals that it is not necessary to reconcile the label difference between users. By leveraging the diversity and training the model in a distributed manner, we can improve the model's performance in personalized humor recognition.

\section{Conclusions and Future Work}
In this paper, we propose FedHumor - a humorous text recognition model following the federated learning paradigm - to perform personalized humor recognition based on labels from distributed data sources. FedHumor is able to account for the diversity in each person's humor preference in the prediction of perceived humorous content. Through extensive experiments, we compared FedHumor against nine state-of-the-art approaches and evaluated the impact of the federated training approach. The results show that it is able to achieve better generalization performance in humor recognition. To the best of our knowledge, FedHumor is the first federated learning-based personalized humor recognition model.

The conceptualization and evaluation of personalized humor recognition under the FL paradigm has many practical values. In FL, data heterogeneity can sometimes reflect users' preferences in highly subjective tasks such as humor recognition and sarcasm detection \cite{guo-etal-2021-latent}. Through FedHumor, we show that preserving and adapting to data heterogeneity can bring performance gains in the face of diverse user preferences. Through joint training of the text encoder and adapting the predictions for each user, FedHumor can be used to recommend personalized jokes for each user on their edge devices during typing or reading.

Nevertheless, the studies are limited by the availability of suitable real-world datasets. The evaluation setting is designed to cover all kinds of user preferences. However, in practice, we may only be able to collect datasets labeled according to a partial set of user preferences. In addition, the disparity in labels not only comes from people with different humor preferences, but can also result from time-delayed awareness. For example, \cite{mottini2019you} observed that the same joke could be labeled differently even by the same person after some time. This problem relates to the research field of concept shift \cite{lu2018learning}, which has made a lot of progress in adapting the model to changing target functions over time \cite{frias2016online,cano2019bayesian}. Currently, there are no publicly available datasets for studying the concept shift problem in humor perception. In the future work, we will collect the appropriate datasets for studying the problem in which the objective function for the same person changes over time.

\begin{acks}
This work is supported, in part, by Alibaba Group through the Alibaba Innovative Research (AIR) Program and the Alibaba-NTU Singapore Joint Research Institute (JRI), Nanyang Technological University, Singapore; the Nanyang Assistant/Associate Professorships (NAP); The RIE 2020 Advanced Manufacturing and Engineering Programmatic Fund (No. A20G8b0102), Singapore; NTU-SDU-CFAIR (NSC-2019-011); the National Natural Science Foundation of China under Grant NSFC 62106167; the National Research Foundation, Prime Minister’s Office, Singapore through the AI Singapore Programme (AISG2-RP-2020-019), NRF Investigatorship Programme (NRFI Award No. NRF-NRFI05-2019-0002) and NRF Fellowship (NRF-NRFF13-2021-0006). Any opinions, findings and conclusions or recommendations expressed in this material are those of the author(s) and do not reflect the views of National Research Foundation, Singapore.

\end{acks}
  
\bibliographystyle{ACM-Reference-Format}
\bibliography{sample-base}


\begin{thebibliography}{43}


\ifx \showCODEN    \undefined \def \showCODEN     #1{\unskip}     \fi
\ifx \showDOI      \undefined \def \showDOI       #1{#1}\fi
\ifx \showISBNx    \undefined \def \showISBNx     #1{\unskip}     \fi
\ifx \showISBNxiii \undefined \def \showISBNxiii  #1{\unskip}     \fi
\ifx \showISSN     \undefined \def \showISSN      #1{\unskip}     \fi
\ifx \showLCCN     \undefined \def \showLCCN      #1{\unskip}     \fi
\ifx \shownote     \undefined \def \shownote      #1{#1}          \fi
\ifx \showarticletitle \undefined \def \showarticletitle #1{#1}   \fi
\ifx \showURL      \undefined \def \showURL       {\relax}        \fi
\providecommand\bibfield[2]{#2}
\providecommand\bibinfo[2]{#2}
\providecommand\natexlab[1]{#1}
\providecommand\showeprint[2][]{arXiv:#2}

\bibitem[\protect\citeauthoryear{Arivazhagan, Aggarwal, Singh, and
  Choudhary}{Arivazhagan et~al\mbox{.}}{2019}]%
        {arivazhagan2019federated}
\bibfield{author}{\bibinfo{person}{Manoj~Ghuhan Arivazhagan},
  \bibinfo{person}{Vinay Aggarwal}, \bibinfo{person}{Aaditya~Kumar Singh},
  {and} \bibinfo{person}{Sunav Choudhary}.} \bibinfo{year}{2019}\natexlab{}.
\newblock \showarticletitle{Federated Learning with Personalization Layers}.
\newblock \bibinfo{journal}{\emph{CoRR, arXiv:1912.00818}}
  (\bibinfo{year}{2019}).
\newblock


\bibitem[\protect\citeauthoryear{Asad, Moustafa, and Ito}{Asad
  et~al\mbox{.}}{2020}]%
        {asad2020fedopt}
\bibfield{author}{\bibinfo{person}{Muhammad Asad}, \bibinfo{person}{Ahmed
  Moustafa}, {and} \bibinfo{person}{Takayuki Ito}.}
  \bibinfo{year}{2020}\natexlab{}.
\newblock \showarticletitle{FedOpt: towards communication efficiency and
  privacy preservation in federated learning}.
\newblock \bibinfo{journal}{\emph{Applied Sciences}} \bibinfo{volume}{10},
  \bibinfo{number}{8} (\bibinfo{year}{2020}), \bibinfo{pages}{2864}.
\newblock


\bibitem[\protect\citeauthoryear{Aykan and Nal{\c{c}}ac{\i}}{Aykan and
  Nal{\c{c}}ac{\i}}{2018}]%
        {aykan2018assessing}
\bibfield{author}{\bibinfo{person}{Simge Aykan} {and} \bibinfo{person}{Erhan
  Nal{\c{c}}ac{\i}}.} \bibinfo{year}{2018}\natexlab{}.
\newblock \showarticletitle{Assessing Theory of Mind by Humor: The Humor
  Comprehension and Appreciation Test (ToM-HCAT)}.
\newblock \bibinfo{journal}{\emph{Frontiers in psychology}}
  \bibinfo{volume}{9} (\bibinfo{year}{2018}).
\newblock


\bibitem[\protect\citeauthoryear{Binsted, Nijholt, Stock, Strapparava, Ritchie,
  Manurung, Pain, Waller, and O'Mara}{Binsted et~al\mbox{.}}{2006}]%
        {binsted2006computational}
\bibfield{author}{\bibinfo{person}{Kim Binsted}, \bibinfo{person}{Anton
  Nijholt}, \bibinfo{person}{Oliviero Stock}, \bibinfo{person}{Carlo
  Strapparava}, \bibinfo{person}{G Ritchie}, \bibinfo{person}{R Manurung},
  \bibinfo{person}{H Pain}, \bibinfo{person}{Annalu Waller}, {and}
  \bibinfo{person}{D O'Mara}.} \bibinfo{year}{2006}\natexlab{}.
\newblock \showarticletitle{Computational humor}.
\newblock \bibinfo{journal}{\emph{IEEE Intelligent Systems}}
  \bibinfo{volume}{21}, \bibinfo{number}{2} (\bibinfo{year}{2006}),
  \bibinfo{pages}{59--69}.
\newblock


\bibitem[\protect\citeauthoryear{Cano, G{\'o}mez-Olmedo, and Moral}{Cano
  et~al\mbox{.}}{2019}]%
        {cano2019bayesian}
\bibfield{author}{\bibinfo{person}{Andr{\'e}s Cano}, \bibinfo{person}{Manuel
  G{\'o}mez-Olmedo}, {and} \bibinfo{person}{Seraf{\'\i}n Moral}.}
  \bibinfo{year}{2019}\natexlab{}.
\newblock \showarticletitle{A Bayesian approach to abrupt concept drift}.
\newblock \bibinfo{journal}{\emph{Knowledge-Based Systems}}
  \bibinfo{volume}{185} (\bibinfo{year}{2019}), \bibinfo{pages}{104909}.
\newblock


\bibitem[\protect\citeauthoryear{Chen and Lee}{Chen and Lee}{2017}]%
        {chen2017convolutional}
\bibfield{author}{\bibinfo{person}{Lei Chen} {and} \bibinfo{person}{Chong~MIn
  Lee}.} \bibinfo{year}{2017}\natexlab{}.
\newblock \showarticletitle{Convolutional neural network for humor
  recognition}.
\newblock \bibinfo{journal}{\emph{CoRR, arXiv:1702.02584}}
  (\bibinfo{year}{2017}).
\newblock


\bibitem[\protect\citeauthoryear{Chen and Soo}{Chen and Soo}{2018}]%
        {chen2018humor}
\bibfield{author}{\bibinfo{person}{Peng-Yu Chen} {and} \bibinfo{person}{Von-Wun
  Soo}.} \bibinfo{year}{2018}\natexlab{}.
\newblock \showarticletitle{Humor recognition using deep learning}. In
  \bibinfo{booktitle}{\emph{NAACL}}. \bibinfo{pages}{113--117}.
\newblock


\bibitem[\protect\citeauthoryear{Chen, Yang, Qin, Yu, Chen, and Shen}{Chen
  et~al\mbox{.}}{2020}]%
        {chen2020focus}
\bibfield{author}{\bibinfo{person}{Yiqiang Chen}, \bibinfo{person}{Xiaodong
  Yang}, \bibinfo{person}{Xin Qin}, \bibinfo{person}{Han Yu},
  \bibinfo{person}{Biao Chen}, {and} \bibinfo{person}{Zhiqi Shen}.}
  \bibinfo{year}{2020}\natexlab{}.
\newblock \showarticletitle{FOCUS: Dealing with Label Quality Disparity in
  Federated Learning}.
\newblock \bibinfo{journal}{\emph{arXiv preprint arXiv:2001.11359}}
  (\bibinfo{year}{2020}).
\newblock


\bibitem[\protect\citeauthoryear{Devlin, Chang, Lee, and Toutanova}{Devlin
  et~al\mbox{.}}{2019}]%
        {devlin-etal-2019-bert}
\bibfield{author}{\bibinfo{person}{Jacob Devlin}, \bibinfo{person}{Ming-Wei
  Chang}, \bibinfo{person}{Kenton Lee}, {and} \bibinfo{person}{Kristina
  Toutanova}.} \bibinfo{year}{2019}\natexlab{}.
\newblock \showarticletitle{Bert: Pre-training of deep bidirectional
  transformers for language understanding}. \bibinfo{publisher}{Association for
  Computational Linguistics}, \bibinfo{address}{Minneapolis, Minnesota},
  \bibinfo{pages}{4171--4186}.
\newblock
\urldef\tempurl%
\url{https://doi.org/10.18653/v1/N19-1423}
\showDOI{\tempurl}


\bibitem[\protect\citeauthoryear{Fallah, Mokhtari, and Ozdaglar}{Fallah
  et~al\mbox{.}}{2020}]%
        {fallah2020personalized}
\bibfield{author}{\bibinfo{person}{Alireza Fallah}, \bibinfo{person}{Aryan
  Mokhtari}, {and} \bibinfo{person}{Asuman~E Ozdaglar}.}
  \bibinfo{year}{2020}\natexlab{}.
\newblock \showarticletitle{Personalized Federated Learning with Theoretical
  Guarantees: A Model-Agnostic Meta-Learning Approach.}. In
  \bibinfo{booktitle}{\emph{NeurIPS}}.
\newblock


\bibitem[\protect\citeauthoryear{Fr{\'\i}as-Blanco, del Campo-Avila,
  Ramos-Jim{\'e}nez, Carvalho, Ortiz-D{\'\i}az, and
  Morales-Bueno}{Fr{\'\i}as-Blanco et~al\mbox{.}}{2016}]%
        {frias2016online}
\bibfield{author}{\bibinfo{person}{Isvani Fr{\'\i}as-Blanco},
  \bibinfo{person}{Jose del Campo-Avila}, \bibinfo{person}{Gonzalo
  Ramos-Jim{\'e}nez}, \bibinfo{person}{Andre~CPLF Carvalho},
  \bibinfo{person}{Agust{\'\i}n Ortiz-D{\'\i}az}, {and} \bibinfo{person}{Rafael
  Morales-Bueno}.} \bibinfo{year}{2016}\natexlab{}.
\newblock \showarticletitle{Online adaptive decision trees based on
  concentration inequalities}.
\newblock \bibinfo{journal}{\emph{Knowledge-Based Systems}}
  \bibinfo{volume}{104} (\bibinfo{year}{2016}), \bibinfo{pages}{179--194}.
\newblock


\bibitem[\protect\citeauthoryear{Geirhos, Jacobsen, Michaelis, Zemel, Brendel,
  Bethge, and Wichmann}{Geirhos et~al\mbox{.}}{2020}]%
        {geirhos2020shortcut}
\bibfield{author}{\bibinfo{person}{Robert Geirhos},
  \bibinfo{person}{J{\"o}rn-Henrik Jacobsen}, \bibinfo{person}{Claudio
  Michaelis}, \bibinfo{person}{Richard Zemel}, \bibinfo{person}{Wieland
  Brendel}, \bibinfo{person}{Matthias Bethge}, {and} \bibinfo{person}{Felix~A
  Wichmann}.} \bibinfo{year}{2020}\natexlab{}.
\newblock \showarticletitle{Shortcut learning in deep neural networks}.
\newblock \bibinfo{journal}{\emph{Nature Machine Intelligence}}
  \bibinfo{volume}{2}, \bibinfo{number}{11} (\bibinfo{year}{2020}),
  \bibinfo{pages}{665--673}.
\newblock


\bibitem[\protect\citeauthoryear{Gruner}{Gruner}{1997}]%
        {Gruner:1997}
\bibfield{author}{\bibinfo{person}{Charles~R. Gruner}.}
  \bibinfo{year}{1997}\natexlab{}.
\newblock \showarticletitle{The Game of Humor: A Comprehensive Theory of Why We
  Laugh}.
\newblock \bibinfo{journal}{\emph{Transaction Publishers}}
  (\bibinfo{year}{1997}).
\newblock


\bibitem[\protect\citeauthoryear{Guo, Li, Yu, and Miao}{Guo
  et~al\mbox{.}}{2021}]%
        {guo-etal-2021-latent}
\bibfield{author}{\bibinfo{person}{Xu Guo}, \bibinfo{person}{Boyang Li},
  \bibinfo{person}{Han Yu}, {and} \bibinfo{person}{Chunyan Miao}.}
  \bibinfo{year}{2021}\natexlab{}.
\newblock \showarticletitle{Latent-Optimized Adversarial Neural Transfer for
  Sarcasm Detection}. In \bibinfo{booktitle}{\emph{Proceedings of the 2021
  Conference of the North American Chapter of the Association for Computational
  Linguistics: Human Language Technologies}}. \bibinfo{publisher}{Association
  for Computational Linguistics}, \bibinfo{address}{Online},
  \bibinfo{pages}{5394--5407}.
\newblock
\urldef\tempurl%
\url{https://doi.org/10.18653/v1/2021.naacl-main.425}
\showDOI{\tempurl}


\bibitem[\protect\citeauthoryear{Heintz and Ruch}{Heintz and Ruch}{2019}]%
        {heintz2019four}
\bibfield{author}{\bibinfo{person}{Sonja Heintz} {and}
  \bibinfo{person}{Willibald Ruch}.} \bibinfo{year}{2019}\natexlab{}.
\newblock \showarticletitle{From four to nine styles: An update on individual
  differences in humor}.
\newblock \bibinfo{journal}{\emph{Personality and Individual Differences}}
  \bibinfo{volume}{141} (\bibinfo{year}{2019}), \bibinfo{pages}{7--12}.
\newblock


\bibitem[\protect\citeauthoryear{Hossain, Krumm, and Gamon}{Hossain
  et~al\mbox{.}}{2019}]%
        {hossain2019president}
\bibfield{author}{\bibinfo{person}{Nabil Hossain}, \bibinfo{person}{John
  Krumm}, {and} \bibinfo{person}{Michael Gamon}.}
  \bibinfo{year}{2019}\natexlab{}.
\newblock \showarticletitle{``{President Vows to Cut Hair}'': Dataset and
  Analysis of Creative Text Editing for Humorous Headlines}.
\newblock \bibinfo{journal}{\emph{CoRR, arXiv:1906.00274}}
  (\bibinfo{year}{2019}).
\newblock


\bibitem[\protect\citeauthoryear{Kairouz, McMahan, Avent, Bellet, Bennis,
  Bhagoji, Bonawitz, Charles, Cormode, Cummings, et~al\mbox{.}}{Kairouz
  et~al\mbox{.}}{2019}]%
        {kairouz2019advances}
\bibfield{author}{\bibinfo{person}{Peter Kairouz}, \bibinfo{person}{H~Brendan
  McMahan}, \bibinfo{person}{Brendan Avent}, \bibinfo{person}{Aur{\'e}lien
  Bellet}, \bibinfo{person}{Mehdi Bennis}, \bibinfo{person}{Arjun~Nitin
  Bhagoji}, \bibinfo{person}{Keith Bonawitz}, \bibinfo{person}{Zachary
  Charles}, \bibinfo{person}{Graham Cormode}, \bibinfo{person}{Rachel
  Cummings}, {et~al\mbox{.}}} \bibinfo{year}{2019}\natexlab{}.
\newblock \showarticletitle{Advances and open problems in federated learning}.
\newblock \bibinfo{journal}{\emph{arXiv preprint arXiv:1912.04977}}
  (\bibinfo{year}{2019}).
\newblock


\bibitem[\protect\citeauthoryear{Lan, Chen, Goodman, Gimpel, Sharma, and
  Soricut}{Lan et~al\mbox{.}}{2019}]%
        {lan2019albert}
\bibfield{author}{\bibinfo{person}{Zhenzhong Lan}, \bibinfo{person}{Mingda
  Chen}, \bibinfo{person}{Sebastian Goodman}, \bibinfo{person}{Kevin Gimpel},
  \bibinfo{person}{Piyush Sharma}, {and} \bibinfo{person}{Radu Soricut}.}
  \bibinfo{year}{2019}\natexlab{}.
\newblock \showarticletitle{Albert: A lite bert for self-supervised learning of
  language representations}.
\newblock \bibinfo{journal}{\emph{CoRR, arXiv:1909.11942}}
  (\bibinfo{year}{2019}).
\newblock


\bibitem[\protect\citeauthoryear{Li, Hu, Beirami, and Smith}{Li
  et~al\mbox{.}}{2020}]%
        {li2020federated}
\bibfield{author}{\bibinfo{person}{Tian Li}, \bibinfo{person}{Shengyuan Hu},
  \bibinfo{person}{Ahmad Beirami}, {and} \bibinfo{person}{Virginia Smith}.}
  \bibinfo{year}{2020}\natexlab{}.
\newblock \showarticletitle{Federated multi-task learning for competing
  constraints}.
\newblock \bibinfo{journal}{\emph{arXiv preprint arXiv:2012.04221}}
  (\bibinfo{year}{2020}).
\newblock


\bibitem[\protect\citeauthoryear{Li, Sahu, Zaheer, Sanjabi, Talwalkar, and
  Smith}{Li et~al\mbox{.}}{2018}]%
        {li2018federated}
\bibfield{author}{\bibinfo{person}{Tian Li}, \bibinfo{person}{Anit~Kumar Sahu},
  \bibinfo{person}{Manzil Zaheer}, \bibinfo{person}{Maziar Sanjabi},
  \bibinfo{person}{Ameet Talwalkar}, {and} \bibinfo{person}{Virginia Smith}.}
  \bibinfo{year}{2018}\natexlab{}.
\newblock \showarticletitle{Federated optimization in heterogeneous networks}.
\newblock \bibinfo{journal}{\emph{arXiv preprint arXiv:1812.06127}}
  (\bibinfo{year}{2018}).
\newblock


\bibitem[\protect\citeauthoryear{Liu, Zhang, and Song}{Liu
  et~al\mbox{.}}{2018}]%
        {liu2018exploiting}
\bibfield{author}{\bibinfo{person}{Lizhen Liu}, \bibinfo{person}{Donghai
  Zhang}, {and} \bibinfo{person}{Wei Song}.} \bibinfo{year}{2018}\natexlab{}.
\newblock \showarticletitle{Exploiting syntactic structures for humor
  recognition}. In \bibinfo{booktitle}{\emph{Proceedings of the 27th
  International Conference on Computational Linguistics}}.
  \bibinfo{pages}{1875--1883}.
\newblock


\bibitem[\protect\citeauthoryear{Lu, Liu, Dong, Gu, Gama, and Zhang}{Lu
  et~al\mbox{.}}{2018}]%
        {lu2018learning}
\bibfield{author}{\bibinfo{person}{Jie Lu}, \bibinfo{person}{Anjin Liu},
  \bibinfo{person}{Fan Dong}, \bibinfo{person}{Feng Gu}, \bibinfo{person}{Joao
  Gama}, {and} \bibinfo{person}{Guangquan Zhang}.}
  \bibinfo{year}{2018}\natexlab{}.
\newblock \showarticletitle{Learning under concept drift: A review}.
\newblock \bibinfo{journal}{\emph{IEEE Transactions on Knowledge and Data
  Engineering}} \bibinfo{volume}{31}, \bibinfo{number}{12}
  (\bibinfo{year}{2018}), \bibinfo{pages}{2346--2363}.
\newblock


\bibitem[\protect\citeauthoryear{Lyu, Yu, Nandakumar, Li, Ma, Jin, Yu, and
  Ng}{Lyu et~al\mbox{.}}{2020}]%
        {lyu2020towards}
\bibfield{author}{\bibinfo{person}{Lingjuan Lyu}, \bibinfo{person}{Jiangshan
  Yu}, \bibinfo{person}{Karthik Nandakumar}, \bibinfo{person}{Yitong Li},
  \bibinfo{person}{Xingjun Ma}, \bibinfo{person}{Jiong Jin},
  \bibinfo{person}{Han Yu}, {and} \bibinfo{person}{Kee~Siong Ng}.}
  \bibinfo{year}{2020}\natexlab{}.
\newblock \showarticletitle{Towards Fair and Privacy-Preserving Federated Deep
  Models}.
\newblock \bibinfo{journal}{\emph{IEEE Transactions on Parallel and Distributed
  Systems}} \bibinfo{volume}{31}, \bibinfo{number}{11} (\bibinfo{year}{2020}),
  \bibinfo{pages}{2524--2541}.
\newblock


\bibitem[\protect\citeauthoryear{Mao and Liu}{Mao and Liu}{2019}]%
        {mao2019bert}
\bibfield{author}{\bibinfo{person}{Jihang Mao} {and} \bibinfo{person}{Wanli
  Liu}.} \bibinfo{year}{2019}\natexlab{}.
\newblock \showarticletitle{A BERT-based Approach for Automatic Humor Detection
  and Scoring}. In \bibinfo{booktitle}{\emph{IberLEF}}.
\newblock


\bibitem[\protect\citeauthoryear{Martin, Puhlik-Doris, Larsen, Gray, and
  Weir}{Martin et~al\mbox{.}}{2003}]%
        {martin2003individual}
\bibfield{author}{\bibinfo{person}{Rod~A Martin}, \bibinfo{person}{Patricia
  Puhlik-Doris}, \bibinfo{person}{Gwen Larsen}, \bibinfo{person}{Jeanette
  Gray}, {and} \bibinfo{person}{Kelly Weir}.} \bibinfo{year}{2003}\natexlab{}.
\newblock \showarticletitle{Individual differences in uses of humor and their
  relation to psychological well-being: Development of the Humor Styles
  Questionnaire}.
\newblock \bibinfo{journal}{\emph{Journal of research in personality}}
  \bibinfo{volume}{37}, \bibinfo{number}{1} (\bibinfo{year}{2003}),
  \bibinfo{pages}{48--75}.
\newblock


\bibitem[\protect\citeauthoryear{McMahan, Moore, Ramage, Hampson, and
  y~Arcas}{McMahan et~al\mbox{.}}{2017}]%
        {pmlr-v54-mcmahan17a}
\bibfield{author}{\bibinfo{person}{Brendan McMahan}, \bibinfo{person}{Eider
  Moore}, \bibinfo{person}{Daniel Ramage}, \bibinfo{person}{Seth Hampson},
  {and} \bibinfo{person}{Blaise~Aguera y Arcas}.}
  \bibinfo{year}{2017}\natexlab{}.
\newblock \showarticletitle{{Communication-Efficient Learning of Deep Networks
  from Decentralized Data}}. In \bibinfo{booktitle}{\emph{AISTATS}}.
  \bibinfo{pages}{1273--1282}.
\newblock


\bibitem[\protect\citeauthoryear{Mihalcea and Pulman}{Mihalcea and
  Pulman}{2007}]%
        {mihalcea2007characterizing}
\bibfield{author}{\bibinfo{person}{Rada Mihalcea} {and}
  \bibinfo{person}{Stephen Pulman}.} \bibinfo{year}{2007}\natexlab{}.
\newblock \showarticletitle{Characterizing humour: An exploration of features
  in humorous texts}. In \bibinfo{booktitle}{\emph{International Conference on
  Intelligent Text Processing and Computational Linguistics}}. Springer,
  \bibinfo{pages}{337--347}.
\newblock


\bibitem[\protect\citeauthoryear{Mihalcea and Strapparava}{Mihalcea and
  Strapparava}{2005}]%
        {mihalcea2005making}
\bibfield{author}{\bibinfo{person}{Rada Mihalcea} {and} \bibinfo{person}{Carlo
  Strapparava}.} \bibinfo{year}{2005}\natexlab{}.
\newblock \showarticletitle{Making computers laugh: Investigations in automatic
  humor recognition}. In \bibinfo{booktitle}{\emph{EMNLP}}.
  \bibinfo{pages}{531--538}.
\newblock


\bibitem[\protect\citeauthoryear{Mikolov, Sutskever, Chen, Corrado, and
  Dean}{Mikolov et~al\mbox{.}}{2013}]%
        {mikolov2013distributed}
\bibfield{author}{\bibinfo{person}{Tomas Mikolov}, \bibinfo{person}{Ilya
  Sutskever}, \bibinfo{person}{Kai Chen}, \bibinfo{person}{Greg~S Corrado},
  {and} \bibinfo{person}{Jeff Dean}.} \bibinfo{year}{2013}\natexlab{}.
\newblock \showarticletitle{Distributed representations of words and phrases
  and their compositionality}. In \bibinfo{booktitle}{\emph{NeurIPS}}.
  \bibinfo{pages}{3111--3119}.
\newblock


\bibitem[\protect\citeauthoryear{Miller, Hempelmann, and Gurevych}{Miller
  et~al\mbox{.}}{2017}]%
        {miller2017semeval}
\bibfield{author}{\bibinfo{person}{Tristan Miller},
  \bibinfo{person}{Christian~F Hempelmann}, {and} \bibinfo{person}{Iryna
  Gurevych}.} \bibinfo{year}{2017}\natexlab{}.
\newblock \showarticletitle{Semeval-2017 task 7: Detection and interpretation
  of english puns}. In \bibinfo{booktitle}{\emph{Proceedings of the 11th
  International Workshop on Semantic Evaluation (SemEval-2017)}}.
  \bibinfo{pages}{58--68}.
\newblock


\bibitem[\protect\citeauthoryear{Mottini and Chowdhury}{Mottini and
  Chowdhury}{2019}]%
        {mottini2019you}
\bibfield{author}{\bibinfo{person}{Alejandro Mottini} {and}
  \bibinfo{person}{Amber~Roy Chowdhury}.} \bibinfo{year}{2019}\natexlab{}.
\newblock \showarticletitle{What Do You Mean I'm Funny? Personalizing the Joke
  Skill of a Voice-Controlled Virtual Assistant}.
\newblock \bibinfo{journal}{\emph{arXiv preprint arXiv:1912.03234}}
  (\bibinfo{year}{2019}).
\newblock


\bibitem[\protect\citeauthoryear{Potash, Romanov, and Rumshisky}{Potash
  et~al\mbox{.}}{2017}]%
        {potash-etal-2017-semeval}
\bibfield{author}{\bibinfo{person}{Peter Potash}, \bibinfo{person}{Alexey
  Romanov}, {and} \bibinfo{person}{Anna Rumshisky}.}
  \bibinfo{year}{2017}\natexlab{}.
\newblock \showarticletitle{{S}em{E}val-2017 Task 6: {\#}{H}ashtag{W}ars:
  Learning a Sense of Humor}. In \bibinfo{booktitle}{\emph{Proceedings of the
  11th International Workshop on Semantic Evaluation ({S}em{E}val-2017)}}.
  \bibinfo{publisher}{Association for Computational Linguistics},
  \bibinfo{address}{Vancouver, Canada}, \bibinfo{pages}{49--57}.
\newblock
\urldef\tempurl%
\url{https://doi.org/10.18653/v1/S17-2004}
\showDOI{\tempurl}


\bibitem[\protect\citeauthoryear{Ramaswamy, Mathews, Rao, and
  Beaufays}{Ramaswamy et~al\mbox{.}}{2019}]%
        {ramaswamy2019federated}
\bibfield{author}{\bibinfo{person}{Swaroop Ramaswamy}, \bibinfo{person}{Rajiv
  Mathews}, \bibinfo{person}{Kanishka Rao}, {and}
  \bibinfo{person}{Fran{\c{c}}oise Beaufays}.} \bibinfo{year}{2019}\natexlab{}.
\newblock \showarticletitle{Federated learning for emoji prediction in a mobile
  keyboard}.
\newblock \bibinfo{journal}{\emph{CoRR, arXiv:1906.04329}}
  (\bibinfo{year}{2019}).
\newblock


\bibitem[\protect\citeauthoryear{Shultz}{Shultz}{1976}]%
        {Shultz:1976}
\bibfield{author}{\bibinfo{person}{Thomas~R Shultz}.}
  \bibinfo{year}{1976}\natexlab{}.
\newblock \showarticletitle{A cognitive-developmental analysis of humour.}
\newblock  (\bibinfo{year}{1976}).
\newblock


\bibitem[\protect\citeauthoryear{Smith, Chiang, Sanjabi, and Talwalkar}{Smith
  et~al\mbox{.}}{2017}]%
        {smith2017federated}
\bibfield{author}{\bibinfo{person}{Virginia Smith}, \bibinfo{person}{Chao-Kai
  Chiang}, \bibinfo{person}{Maziar Sanjabi}, {and} \bibinfo{person}{Ameet
  Talwalkar}.} \bibinfo{year}{2017}\natexlab{}.
\newblock \showarticletitle{Federated multi-task learning}.
\newblock \bibinfo{journal}{\emph{arXiv preprint arXiv:1705.10467}}
  (\bibinfo{year}{2017}).
\newblock


\bibitem[\protect\citeauthoryear{Taylor and Mazlack}{Taylor and
  Mazlack}{2004}]%
        {taylor2004computationally}
\bibfield{author}{\bibinfo{person}{Julia~M Taylor} {and}
  \bibinfo{person}{Lawrence~J Mazlack}.} \bibinfo{year}{2004}\natexlab{}.
\newblock \showarticletitle{Computationally recognizing wordplay in jokes}. In
  \bibinfo{booktitle}{\emph{Proceedings of the Annual Meeting of the Cognitive
  Science Society}}, Vol.~\bibinfo{volume}{26}.
\newblock


\bibitem[\protect\citeauthoryear{Veale}{Veale}{2012}]%
        {veale2012exploding}
\bibfield{author}{\bibinfo{person}{Tony Veale}.}
  \bibinfo{year}{2012}\natexlab{}.
\newblock \bibinfo{booktitle}{\emph{Exploding the creativity myth: The
  computational foundations of linguistic creativity}}.
\newblock \bibinfo{publisher}{A\&C Black}.
\newblock


\bibitem[\protect\citeauthoryear{Wimer and Beins}{Wimer and Beins}{2008}]%
        {wimer2008expectations}
\bibfield{author}{\bibinfo{person}{David~J Wimer} {and}
  \bibinfo{person}{Bernard~C Beins}.} \bibinfo{year}{2008}\natexlab{}.
\newblock \showarticletitle{Expectations and perceived humor}.
\newblock \bibinfo{journal}{\emph{Humor}} \bibinfo{volume}{21},
  \bibinfo{number}{3} (\bibinfo{year}{2008}), \bibinfo{pages}{347--363}.
\newblock


\bibitem[\protect\citeauthoryear{Yang, Lavie, Dyer, and Hovy}{Yang
  et~al\mbox{.}}{2015}]%
        {yang2015humor}
\bibfield{author}{\bibinfo{person}{Diyi Yang}, \bibinfo{person}{Alon Lavie},
  \bibinfo{person}{Chris Dyer}, {and} \bibinfo{person}{Eduard Hovy}.}
  \bibinfo{year}{2015}\natexlab{}.
\newblock \showarticletitle{Humor recognition and humor anchor extraction}. In
  \bibinfo{booktitle}{\emph{EMNLP}}. \bibinfo{pages}{2367--2376}.
\newblock


\bibitem[\protect\citeauthoryear{Yang, Liu, Chen, and Tong}{Yang
  et~al\mbox{.}}{2019a}]%
        {yang2019federated}
\bibfield{author}{\bibinfo{person}{Qiang Yang}, \bibinfo{person}{Yang Liu},
  \bibinfo{person}{Tianjian Chen}, {and} \bibinfo{person}{Yongxin Tong}.}
  \bibinfo{year}{2019}\natexlab{a}.
\newblock \showarticletitle{Federated machine learning: Concept and
  applications}.
\newblock \bibinfo{journal}{\emph{ACM Transactions on Intelligent Systems and
  Technology (TIST)}} \bibinfo{volume}{10}, \bibinfo{number}{2}
  (\bibinfo{year}{2019}), \bibinfo{pages}{1--19}.
\newblock


\bibitem[\protect\citeauthoryear{Yang, Liu, Cheng, Kang, Chen, and Yu}{Yang
  et~al\mbox{.}}{2019b}]%
        {FL:2019}
\bibfield{author}{\bibinfo{person}{Qiang Yang}, \bibinfo{person}{Yang Liu},
  \bibinfo{person}{Yong Cheng}, \bibinfo{person}{Yan Kang},
  \bibinfo{person}{Tianjian Chen}, {and} \bibinfo{person}{Han Yu}.}
  \bibinfo{year}{2019}\natexlab{b}.
\newblock \bibinfo{booktitle}{\emph{Federated Learning}}.
\newblock \bibinfo{publisher}{Morgan \& Claypool Publishers}. 189 pages.
\newblock


\bibitem[\protect\citeauthoryear{Yu, Liu, Liu, Chen, Cong, Weng, Niyato, and
  Yang}{Yu et~al\mbox{.}}{2020}]%
        {yu2020fairness}
\bibfield{author}{\bibinfo{person}{Han Yu}, \bibinfo{person}{Zelei Liu},
  \bibinfo{person}{Yang Liu}, \bibinfo{person}{Tianjian Chen},
  \bibinfo{person}{Mingshu Cong}, \bibinfo{person}{Xi Weng},
  \bibinfo{person}{Dusit Niyato}, {and} \bibinfo{person}{Qiang Yang}.}
  \bibinfo{year}{2020}\natexlab{}.
\newblock \showarticletitle{A fairness-aware incentive scheme for federated
  learning}. In \bibinfo{booktitle}{\emph{Proceedings of the AAAI/ACM
  Conference on AI, Ethics, and Society}}. \bibinfo{pages}{393--399}.
\newblock


\bibitem[\protect\citeauthoryear{Zhang and Liu}{Zhang and Liu}{2014}]%
        {zhang2014recognizing}
\bibfield{author}{\bibinfo{person}{Renxian Zhang} {and} \bibinfo{person}{Naishi
  Liu}.} \bibinfo{year}{2014}\natexlab{}.
\newblock \showarticletitle{Recognizing humor on twitter}. In
  \bibinfo{booktitle}{\emph{CIKM}}. \bibinfo{pages}{889--898}.
\newblock


\end{thebibliography}

%

\end{document}